%% file: AnonymousSubmission2027.tex
\documentclass[letterpaper]{article} 
\usepackage{aaai2027}  
\usepackage[hyphens]{url}  
\usepackage{graphicx} 
\usepackage{natbib}  
\usepackage{caption} 
\usepackage{algorithm}
\usepackage{algorithmic}
\usepackage{makecell}
\usepackage{amssymb}
\usepackage[table]{xcolor}
\usepackage{booktabs}
\usepackage{multirow}
\usepackage{fvextra}
\usepackage{adjustbox}
\usepackage{ragged2e}
\usepackage{booktabs,tabularx}
\usepackage{subcaption}
\usepackage{enumitem}
\usepackage{newfloat}
\usepackage{listings}
\DeclareCaptionStyle{ruled}{labelfont=normalfont,labelsep=colon,strut=off} 
\floatstyle{ruled}
\newfloat{listing}{tb}{lst}{}
\floatname{listing}{Listing}

\title{Beyond Function Calling: Benchmarking Tool-Using Agents under Tool-Environment Unreliability}
\author{
    Yang Tian\textsuperscript{\rm 1}\equalcontrib,
    Zhengpeng Shi\textsuperscript{\rm 1}\equalcontrib,
    Yu Zhou\textsuperscript{\rm 2},
    Bo Zhao
}
\affiliations{
    School of AI, Shanghai Jiao Tong University\textsuperscript{\rm 1}\\
    School of Cyber Science and Engineering, Southeast University\textsuperscript{\rm 2}

}

\makeatletter
\def\copyright@text{}
\makeatother
\begin{document}

\maketitle

\begin{abstract}
Large language models are increasingly deployed as agents that solve tasks by interacting with external tool environments. Although recent tool-use benchmarks increasingly cover complex task settings, they still largely assume clean, stable, and trustworthy tool environments, leaving tool-environment unreliability insufficiently examined. We introduce ToolBench-X, a benchmark for evaluating agents under recoverable reliability hazards. ToolBench-X contains executable multi-step tasks across diverse domains and sequential, parallel, and mixed workflows, each paired with deterministic tools and a canonical final answer for automatic evaluation. Starting from clean tool environments, ToolBench-X injects five structured hazard types: Specification Drift, Invocation Error, Execution Failure, Output Drift, and Cross-source Conflict. Crucially, each injected instance remains solvable through at least one valid recovery path, such as retrying, fallback, verification, or cross-checking. Experiments reveal a substantial reliability gap: agents that perform well with reliable tools often fail under recoverable hazards. Further analysis shows that failures are driven less by tool-use volume or inference budget than by limited hazard diagnosis and ineffective recovery. Targeted recovery hints recover many failed tasks, while test-time scaling yields more limited gains. These results suggest that tool-use evaluation should move beyond function-call accuracy toward task completion under unreliable tool environments. The code and data is available at \url{https://github.com/Foreverskyou/ToolBench-X}.
\end{abstract}


\section{Introduction}

\begin{table*}[htbp]
\centering
\scriptsize
\renewcommand{\arraystretch}{1.0}
\resizebox{\textwidth}{!}{
\begin{tabular}{lcccccccccc}
\hline
\textbf{Benchmark} 
& \textbf{\makecell{\#\\Tasks}} 
& \textbf{\makecell{\#\\Tools}} 
& \textbf{\makecell{Specification\\Drift}} 
& \textbf{\makecell{Invocation\\Errors}} 
& \textbf{\makecell{Execution\\Failures}} 
& \textbf{\makecell{Output\\Drift}} 
& \textbf{\makecell{Cross-Source\\Conflict}} 
& \textbf{\makecell{Sequential\\Tool-Use}} 
& \textbf{\makecell{Parallel\\Tool-Use}} 
& \textbf{\makecell{Mixed\\Tool-Use}}\\
\hline

\textbf{ToolBench-X} 
& 1106 
& 4956 
& \textcolor{teal}{\checkmark} 
& \textcolor{teal}{\checkmark} 
& \textcolor{teal}{\checkmark} 
& \textcolor{teal}{\checkmark} 
& \textcolor{teal}{\checkmark} 
& \textcolor{teal}{\checkmark} 
& \textcolor{teal}{\checkmark} 
& \textcolor{teal}{\checkmark} \\

\rowcolor{gray!20}
BFCL v3~\citep{patil2025bfcl} 
& 1,000 
& N/R 
& \textcolor{red}{$\times$} 
& \textcolor{red}{$\times$} 
& \textcolor{red}{$\times$} 
& \textcolor{red}{$\times$} 
& \textcolor{red}{$\times$} 
& \textcolor{red}{$\times$} 
& \textcolor{teal}{\checkmark} 
& \textcolor{red}{$\times$} \\

BFCL v2~\citep{patil2025bfcl} 
& 2,251 
& N/R 
& \textcolor{red}{$\times$} 
& \textcolor{red}{$\times$} 
& \textcolor{red}{$\times$} 
& \textcolor{red}{$\times$} 
& \textcolor{red}{$\times$} 
& \textcolor{red}{$\times$} 
& \textcolor{teal}{\checkmark} 
& \textcolor{red}{$\times$} \\

\rowcolor{gray!20}
BFCL v1~\citep{patil2025bfcl} 
& 2,000 
& N/R  
& \textcolor{red}{$\times$} 
& \textcolor{red}{$\times$} 
& \textcolor{red}{$\times$} 
& \textcolor{red}{$\times$} 
& \textcolor{red}{$\times$} 
& \textcolor{red}{$\times$} 
& \textcolor{teal}{\checkmark} 
& \textcolor{red}{$\times$} \\

ToolBench~\citep{qintoolllm} 
& 126,486 
& \makecell{16,464 APIs} 
& \textcolor{red}{$\times$} 
& \textcolor{red}{$\times$} 
& \textcolor{red}{$\times$} 
& \textcolor{red}{$\times$} 
& \textcolor{red}{$\times$} 
& \textcolor{teal}{\checkmark} 
& \textcolor{red}{$\times$} 
& \textcolor{red}{$\times$} \\

\rowcolor{gray!20}
AnyToolBench~\citep{duanytool} 
& 400 
& $>$16,000 APIs 
& \textcolor{red}{$\times$} 
& \textcolor{red}{$\times$} 
& \textcolor{red}{$\times$} 
& \textcolor{red}{$\times$} 
& \textcolor{red}{$\times$} 
& \textcolor{teal}{\checkmark} 
& \textcolor{red}{$\times$} 
& \textcolor{red}{$\times$} \\

$\tau^2$-bench~\citep{barres2025tau} 
& \makecell{279} 
& \makecell{48} 
& \textcolor{red}{$\times$} 
& \textcolor{red}{$\times$} 
& \textcolor{red}{$\times$} 
& \textcolor{red}{$\times$} 
& \textcolor{red}{$\times$} 
& \textcolor{teal}{\checkmark} 
& \textcolor{red}{$\times$} 
& \textcolor{red}{$\times$} \\

\rowcolor{gray!20}
$\tau$-bench~\citep{yao2024tau} 
& 165 
& 28
& \textcolor{red}{$\times$} 
& \textcolor{red}{$\times$} 
& \textcolor{red}{$\times$} 
& \textcolor{red}{$\times$} 
& \textcolor{red}{$\times$} 
& \textcolor{teal}{\checkmark} 
& \textcolor{red}{$\times$} 
& \textcolor{red}{$\times$} \\

T-EVAL~\citep{chen2024t} 
& 553 
& 15 
& \textcolor{red}{$\times$} 
& \textcolor{red}{$\times$} 
& \textcolor{red}{$\times$} 
& \textcolor{red}{$\times$} 
& \textcolor{red}{$\times$} 
& \textcolor{teal}{\checkmark} 
& \textcolor{red}{$\times$} 
& \textcolor{red}{$\times$} \\

\rowcolor{gray!20}
UltraTool~\citep{huang2024planning} 
& 5,824 
& 2,032 
& \textcolor{red}{$\times$} 
& \textcolor{red}{$\times$} 
& \textcolor{red}{$\times$} 
& \textcolor{red}{$\times$} 
& \textcolor{red}{$\times$} 
& \textcolor{teal}{\checkmark} 
& \textcolor{red}{$\times$} 
& \textcolor{red}{$\times$} \\

AgentNoiseBench~\cite{wang2026agentnoisebench}
& -- 
& -- 
& \textcolor{red}{$\times$} 
& \textcolor{red}{$\times$} 
& \textcolor{teal}{$\checkmark$} 
& \textcolor{teal}{$\checkmark$} 
& \textcolor{red}{$\times$} 
& \textcolor{teal}{\checkmark} 
& \textcolor{red}{$\times$} 
& \textcolor{red}{$\times$} \\
\hline
\end{tabular}
}
\caption{Comparative analysis of ToolBench-X against representative tool-use benchmarks.}
\label{tab:benchmark_reliability_comparison}
\end{table*}

Large language models (LLMs) are advancing at an unprecedented pace, and the development of agents built upon them has emerged as a promising direction in AI research~\cite{deepseekv3,yang2025qwen3}. These agents interact with the real world through diverse tools, thereby unlocking novel opportunities for practical applications. Consequently, establishing robust benchmarks to reliably evaluate the tool-use capabilities of LLMs has become increasingly critical~\cite{huang2024planning,du2024anytool,yao2024tau,guo2025deepseek}.

However, correct function calling is only a necessary condition for reliable tool use. In real-world systems, tools are rarely perfectly specified, perfectly stable, or perfectly trustworthy. API documentation may become stale; expected fields may be renamed or wrapped; services may timeout; returned values may be incomplete, non-canonical, or semantically shifted; and different tools may provide conflicting evidence. Even when a correct answer remains reachable, an agent may fail by trusting a suspicious intermediate result, retrying the wrong call, inventing missing arguments, or finishing before the required evidence is complete. These failures are not captured well by evaluations that assume clean tool contracts and focus primarily on whether the model selected the right function with the right arguments.

This work posits that advancing the evaluation of tool-using agents requires moving beyond mere function invocation: from assessing isolated call correctness to evaluating robust task completion under conditions of structured tool uncertainty. A competent tool agent should not only execute tools correctly but also recognize when tool outputs are unreliable, validate incomplete or contradictory results, recover from runtime errors, and ultimately generate a benchmark-consistent final answer. In other words, evaluating tool use should go beyond asking ``Can the model invoke the tool?'' to also consider ``Can the model accomplish the task when the tool environment is uncertain?''

To investigate this question, we introduce a challenging benchmark for evaluating tool-using agents under structured Reliability Hazards. The benchmark defines executable multi-step tool tasks spanning sequential, parallel, and mixed workflows. Each task is associated with Python tools and a canonical final answer, enabling automatic execution and precise exact-match evaluation. We then convert previously successful baseline tools into uncertainty-injected versions according to a controlled taxonomy of five uncertainty types: specification uncertainty, invocation uncertainty, execution uncertainty, output uncertainty, and cross-source uncertainty. These categories encompass common real-world failure modes, including contract drift, argument ambiguity, service failures, schema or surface-form drift, and conflicting evidence across multiple tools.

A central design principle of our framework is that injected failures should be both disruptive and recoverable. Consequently, we avoid merely corrupting tools to create unsolvable tasks. Instead, each injected task maintains at least one viable recovery path, which may involve retrying, fallback strategies, cross-checking, normalization, or evidence verification. Our experiments indicate that executing tasks in environments subject to reliability hazards is challenging: all evaluated models achieved success rates below 60\%.

We further observe that providing targeted hints when the model encounters difficulty can substantially increase its task completion success rate. Moreover, our failure-triggered, test-time scaling experiments demonstrate that allowing the model to reflect on its failures and engage in additional reasoning rounds also improves performance. This suggests that some recovery capabilities can emerge from extended inference-time computation. Collectively, these findings indicate that many failures are not attributable to inherently unsolvable tasks, but rather to insufficient failure diagnosis, inadequate evidence verification, and limited recovery strategies.

In summary, this work advances the study of tool-using agents along three key dimensions. First, we develop a scalable framework that conceptualizes reliable tool use as the successful completion of tasks under five distinct sources of uncertainty, thereby extending evaluation beyond simple function-call scenarios. Second, we introduce a rigorous benchmark comprising executable tasks, systematically injected uncertainties, and recovery mechanisms, which enables precise assessment of agent robustness in real-world settings. Third, our empirical results demonstrate that current tool-using agents remain highly fragile under runtime uncertainty; however, structured guidance and test-time recovery strategies can substantially enhance task success rates.

\section{Related Work}

\paragraph{Large language models agents.}
Large language models are increasingly studied not only as text generators, but as agents that perceive task context, plan intermediate steps, take actions, observe feedback, and update their behavior over time. This agentic paradigm has been explored across a wide range of domains. For example, WebArena~\cite{zhou2023tianyue} and Mind2Web~\cite{deng2023mind2web} evaluate agents that operate over realistic websites and user instructions; SWE-bench~\cite{jimenez2023swe} evaluates agents on resolving real GitHub issues by editing code and interacting with execution environments; Voyager~\cite{wang2023voyager} studies an embodied lifelong-learning agent in Minecraft; and AgentBench~\cite{liu2023kaiwen} evaluates LLM agents across diverse interactive environments. These works show that modern agents are expected to complete tasks through interaction with external environments rather than by producing a single static response. Across these settings, tool use is a foundational capability. Whether an agent queries a database, calls a web API, executes code, searches documents, edits files, or invokes a domain-specific simulator, it must decide which external operation to perform, provide valid inputs, interpret the returned observation, and incorporate that observation into subsequent reasoning. Tool use therefore acts as an atomic action interface between language models and the external world. Improvements in tool use directly affect broader agent reliability, because higher-level planning and reasoning depend on the correctness and trustworthiness of these low-level interactions.

\paragraph{Benchmarks for tool-using agents.}
Existing benchmarks have made important progress in evaluating tool-use ability. T-EVAL~\cite{chen2024t}, UltraTool~\cite{huang2024planning}, and MetaTool~\cite{huang2024metatool} assess various sub-capabilities of tool-use, but treat tool invocation as a simple question-answering task, which fails to capture the multi-turn interactive nature of the LLM agent loop. On the other hand, BFCL-V1~\cite{patil2025bfcl} and BFCL-V2~\cite{patil2025bfcl} pioneered the evaluation of parallel tool-use but were still limited to single-turn scenarios. BFCL-V3~\cite{patil2025bfcl} introduced multi-turn evaluation and assessed the sequential multi-step capabilities of LLMs. Beyond evaluating whether agents can use tools correctly, assessing their robustness to imperfect or unreliable tool environments is also an important research direction~\cite{wang2026agentnoisebench,xi2026toolgym}. However, prior studies largely focus on shallow tool perturbations, simple tool-call chains, or single-type untrusted tool environments. In contrast, our work systematically evaluates agents under realistic and unexpected tool-environment uncertainties that may emerge throughout a complete, complex task-execution cycle, assessing their ability to complete task progress in unreliable yet recoverable tool environments.

\begin{figure*}[t] 
    \centering
    \includegraphics[width=0.85\textwidth, height=0.3\textheight]{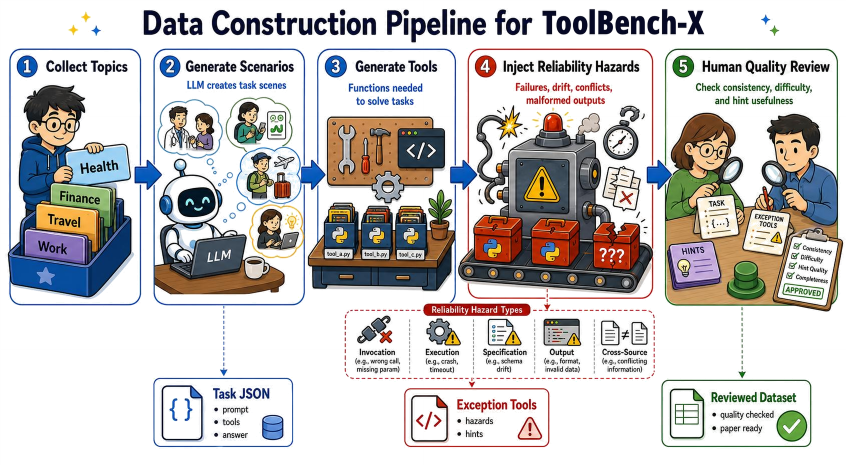} 
    \caption{Benchmark construction pipeline. We first define seven topic categories, generate diverse task scenarios for each topic, synthesize executable tools and canonical answers, inject structured reliability hazards while preserving a valid recovery path, and finally conduct human review to ensure task validity, tool correctness, hazard consistency, and answer reliability.}
    \label{pipeline}
\end{figure*}

\section{ToolBench-X}

\subsection{Problem Formulation}

A tool-using agent can be formalized as a sequential decision-making process. At each step, the agent observes the task context along with previous tool outputs, and selects an action such as invoking a tool with specific arguments, retrying an operation, switching tools, verifying evidence, or producing a final answer. Formally, this can be represented as a Markov Decision Process (MDP):

\[
\mathcal{M}=(\mathcal{S},\mathcal{A},P)
\]

where $\mathcal{S}$ is the state space of environment, $\mathcal{A}$ contains tool calls and the transition kernel $P(s_{t+1} \mid s_t, a_t)$ characterizes how the external tool environment responds to the agent. Existing function-calling benchmarks largely evaluate agents under a clean transition kernel \(P_0\), in which a valid function call \(\texttt{call\_tool}(f,x)\) reliably returns a documented output \(T_f(s_t,x)\). Under this assumption, the primary challenge lies in action correctness: selecting the appropriate function and providing the correct arguments. In practice, however, real-world tool environments often violate this assumption. Runtime failures, outdated schemas, output drift, missing fields, and inconsistencies across tools introduce perturbations, resulting in a modified transition kernel \(P_h\) such that

\[
P_h(s_{t+1}\mid s_t,a_t) \neq P_0(s_{t+1}\mid s_t,a_t)
\]

Since tool-use tasks are inherently multi-step, even localized perturbations can propagate through subsequent argument generation, evidence aggregation, and final answer synthesis. For a fixed policy \(\pi\), the key metric is therefore not only its nominal performance \(V_{P_0}^{\pi}\), but also the degradation \(V_{P_0}^{\pi}-V_{P_h}^{\pi}\) induced by unreliable tools environment. This is crucial, as it better reflects the model’s true capabilities in real-world scenarios.

Furthermore, an environment with unreliable tools is partially observable. The agent cannot directly determine whether a returned field is stale, whether an exception reflects a transient failure, or whether conflicting outputs stem from corrupted sources. As a result, the optimal policy under \(P_h\) may differ substantially from that under \(P_0\), leading to degraded decision quality when the agent relies on incomplete, inconsistent, or unreliable tool observations. This distinction highlights the limitations of conventional function-calling benchmark as a measure of reliable tool-agent performance. A model may perform strongly under \(P_0\) yet exhibit substantial degradation under \(P_h\), because traditional benchmarks rarely capture the reliability hazards present in tool-environment interactions. Our work addresses this gap by explicitly constructing hazards and evaluating both the resulting performance degradation under \(P_h\) and the performance improvements enabled by additional reliability signals. Table~\ref{tab:benchmark_reliability_comparison} shows the comparison of ToolBench-X and other representative tools using benchmarks.

\subsection{Benchmark Construction}

\begin{figure*}[t]
    \centering
    \begin{subfigure}{0.32\textwidth}
        \centering
        \includegraphics[width=\linewidth]{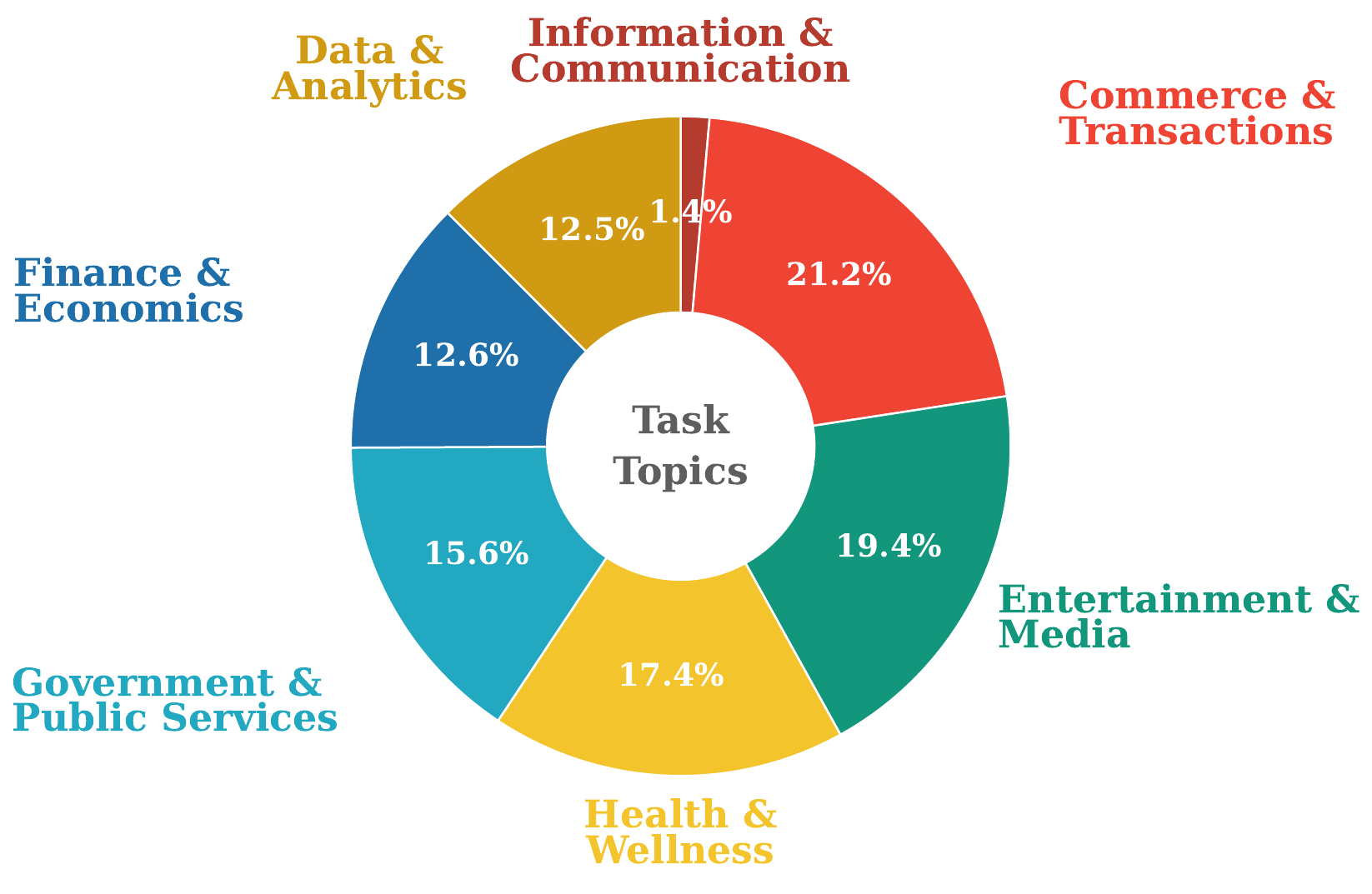}
        \caption{Task domain distribution.}
        \label{fig:topic-donut}
    \end{subfigure}
    \hfill
    \begin{subfigure}{0.32\textwidth}
        \centering
        \includegraphics[width=\linewidth]{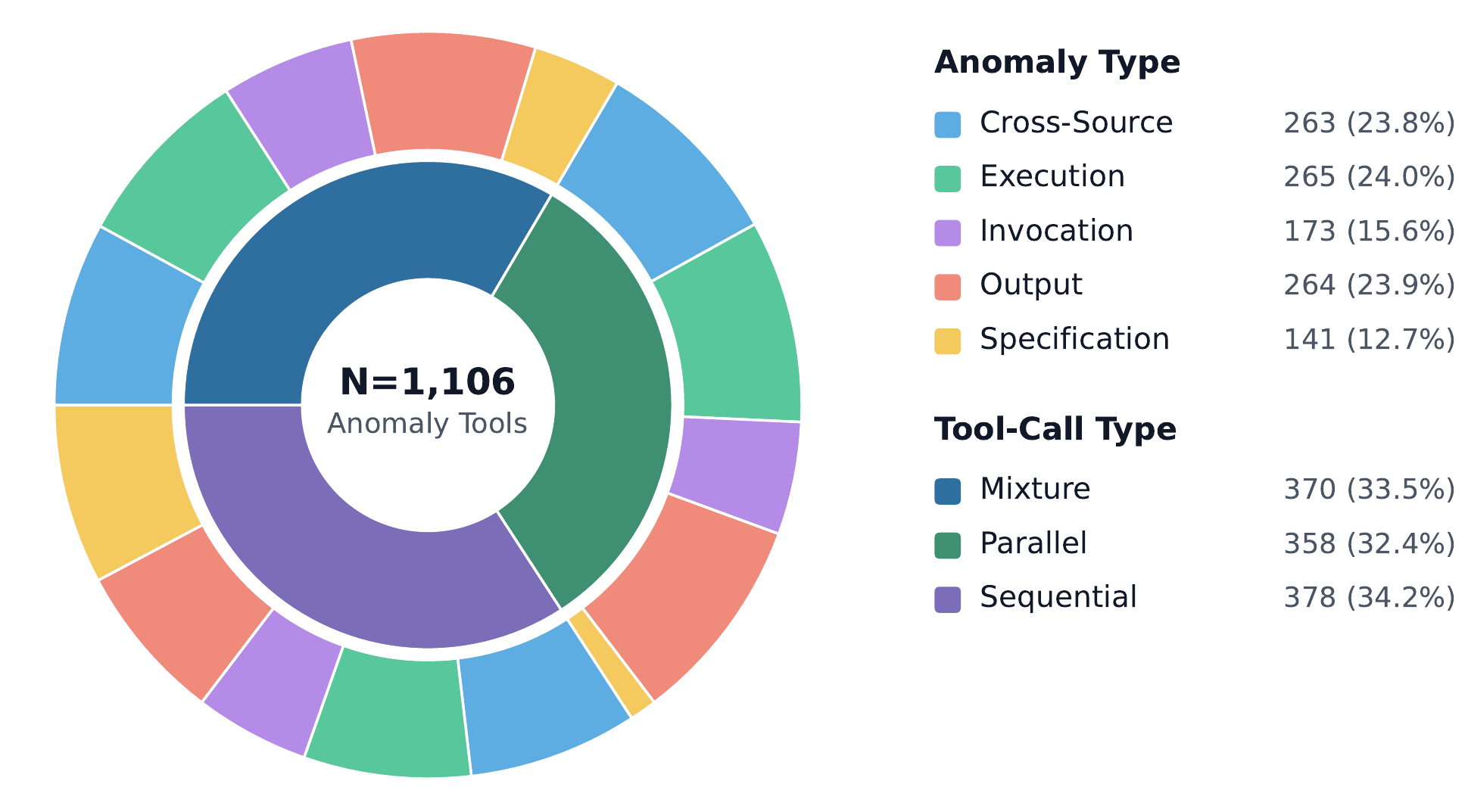}
        \caption{Reliability Hazard types.}
        \label{fig:tools-exception}
    \end{subfigure}
    \hfill
    \begin{subfigure}{0.32\textwidth}
        \centering
        \includegraphics[width=\linewidth]{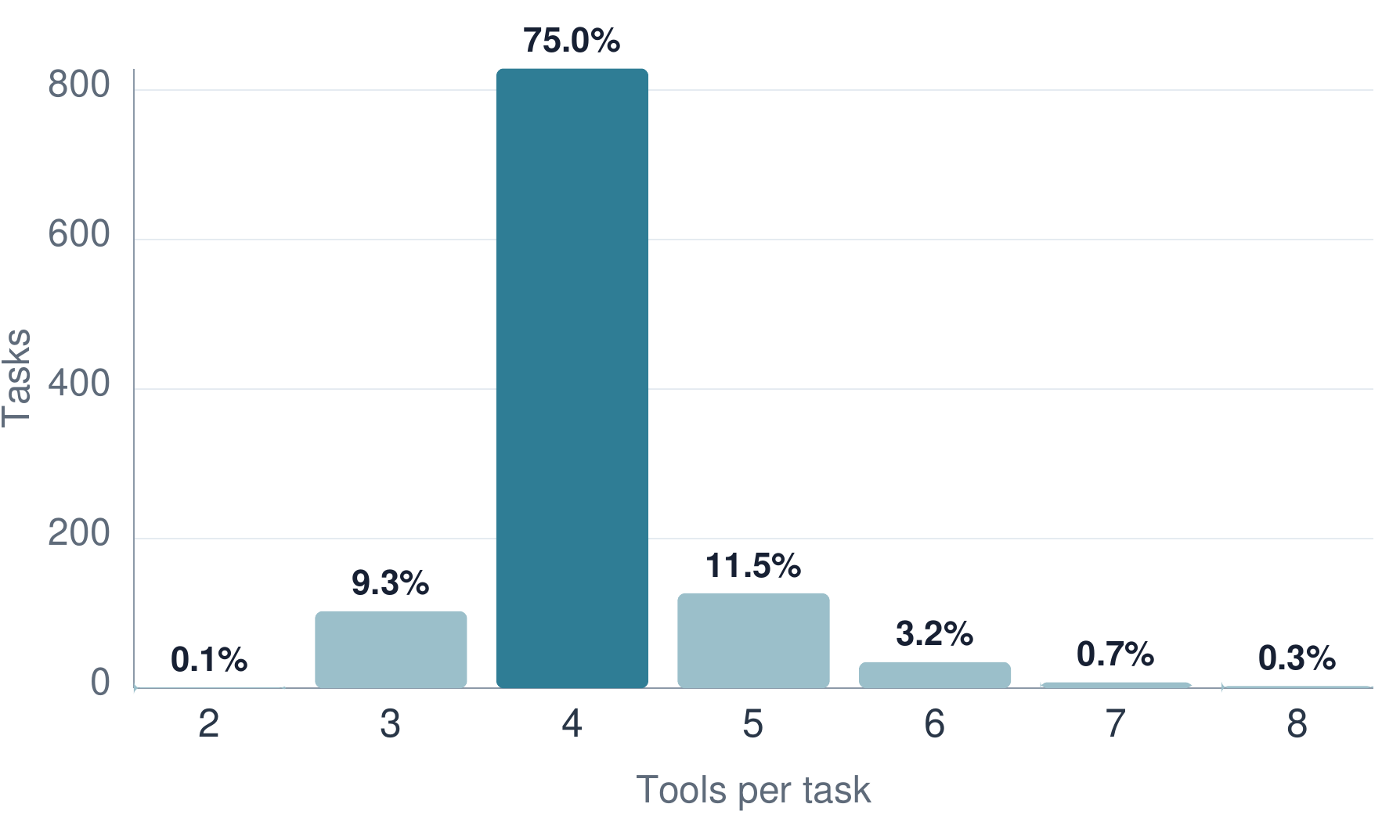}
        \caption{Number of tools per task.}
        \label{fig:tool-count}
    \end{subfigure}

    \caption{Overview of the benchmark distribution across task domains, reliability hazard types, and number of tools per task, showing broad topical coverage, diverse reliability challenges, and varied multi-tool complexity.}
    \label{statistic}
\end{figure*}

We construct the benchmark using a multi-stage pipeline designed to generate diverse, executable, and recoverable tool-use tasks. Our aim is not merely to produce tool-calling problems, but also to introduce controlled reliability hazards that still allow a correct final answer to be reached. The detailed pipeline is illustrated in Fig.~\ref{pipeline}.

\paragraph{Topic and Scenario Generation.}
We begin by defining seven broad topic categories to encompass a wide range of real-world tool-use topics. These topics are designed to capture variations in task intent, information structure, and interaction patterns. For each topic, we prompt a large language model to generate multiple realistic scenarios in which an agent must leverage external tools to fulfill a user request. The resulting scenarios include sequential workflows, parallel workflows, and mixture workflows that incorporate both sequential and parallel processes.

\paragraph{Task and Tool Construction.}
For each scenario, A large language model is employed to produce both detailed task specifications and the Python toolset required for task completion. Each tool is accompanied by a natural language description, input schema, executable implementation, and deterministic output behavior. Additionally, we provide a clean result for each task, in which the task can be successfully solved through correct tool selection, parameter specification, execution, and answer synthesis.

\paragraph{Reliability Hazard Injection.}
Starting from the clean tools, We define five types of tool uncertainty according to where the mismatch between expected and observed tool behavior occurs. In all cases, the ground-truth answer is fixed by the original benchmark expected answer and the deterministic no-injection tool path; exception injection perturbs only the agent’s observable interaction with tools, not the underlying task answer.

Specification Drift occurs when the documented tool contract differs from the runtime contract, such as renamed fields, changed types, altered output shapes, or shifted units. It models stale documentation or API version drift. The correct answer is recovered by mapping the observed runtime behavior back to the intended contract.

Invocation Error occurs when the agent selects the right tool, but the intended call is not faithfully delivered to or interpreted by the tool. Arguments or wrapper-mediated payloads may be dropped, renamed, coerced, defaulted, truncated, or rejected at the tool boundary. It models failures in adapters, middleware, parameter binding, or request construction. The correct answer is recovered by verifying required inputs, reconstructing a valid call, and avoiding downstream use of incomplete or misbound arguments.

Execution Failure occurs after a valid call has reached the tool, but the tool execution is unstable due to timeouts, connection errors, runtime exceptions, or parsing failures. It models unreliable services or execution environments. The correct answer is recovered through bounded retry, fallback, and refusal to finish from partial execution evidence.

Output Drift occurs when the tool result is available but its returned surface form is unstable, such as wrapped values, added units, nested fields, aliases, or non-canonical answer formats. It models formatting and serialization drift. The correct answer is recovered by extracting the underlying value and canonicalizing it to the benchmark-required form.

Cross-source Conflict occurs when the answer depends on multiple tools or evidence branches that may be incomplete, inconsistent, or differently formatted. It models multi-source aggregation and verification. The correct answer is not selected by voting among sources; it is the original benchmark answer, recovered by checking branch completeness, normalizing source outputs, resolving contradictions, and propagating only reconciled evidence.

A key constraint in our construction is recoverability. We do not inject arbitrary corruptions that make the task unsolvable. Instead, every uncertainty-injected instance preserves at least one valid path to the canonical answer. Depending on the uncertainty type, the recovery path may require retrying a failed call, selecting a fallback tool, normalizing a changed output format, checking multiple sources, or verifying suspicious evidence before producing the final answer.

\paragraph{Human Validation.}
Finally, all generated instances undergo rigorous human review. This process verifies that each task is well-posed, the associated tools are executable, the baseline answer is correct, the injected hazard aligns with its intended uncertainty category, and the recovery path is valid. Instances exhibiting ambiguous instructions, multiple plausible final answers, irrecoverable failures, inconsistent tool behavior, or answer leakage through the prompt are either revised or removed. This human-in-the-loop validation ensures that benchmark failures reflect genuine limitations of agent reliability rather than artifacts arising from dataset construction.

\paragraph{Statistics of ToolBench-X}

\begin{table*}[t]
\centering

\resizebox{\textwidth}{!}{%
\begin{tabular}{lccc ccccc c}
\toprule
\multirow{2}{*}{\textbf{Models}} 
& \multicolumn{3}{c}{\textbf{Task Type}} 
& \multicolumn{5}{c}{\textbf{Exception Type}} 
& \multirow{2}{*}{\textbf{Overall}} \\
\cmidrule(lr){2-4} \cmidrule(lr){5-9}
& Parallel & Sequential & Mixture 
& Cross-Source & Execution & Invocation & Output & Specification 
& \\
\midrule
DeepSeek-V4-Pro~\cite{deepseekai2026deepseekv4}   & 0.469 & 0.397 & 0.411 & 0.335 & 0.355 & 0.272 & 0.663 & 0.468 & 0.425 \\
GPT-5.4~\cite{singh2025openai}       & 0.472 & \textbf{0.450} & 0.438 & 0.350 & 0.358 & 0.283 & \underline{0.727} & \textbf{0.518} & \underline{0.453} \\
Doubao-Seed-2.0-Lite~\cite{bytedance2026doubaoseed20lite}        & \textbf{0.587} & \underline{0.439} & \textbf{0.516} & \textbf{0.460} & \textbf{0.457} & \textbf{0.353} & \textbf{0.750} & 0.468 & \textbf{0.513} \\
Claude-Sonnet-4.6~\cite{anthropic2026claudesonnet46systemcard}        & 0.425 & 0.402 & 0.405 & 0.312 & 0.332 & 0.272 & 0.655 & 0.454 & 0.410 \\
Gemini-3.1-Flash~\cite{google2026geminimodels}        & \underline{0.528} & 0.339 & 0.386 & 0.354 & 0.396 & 0.301 & 0.572 & 0.418 & 0.416 \\
GLM-5.1~\cite{zeng2026glm}       & 0.511 & 0.357 & 0.395 & \underline{0.373} & 0.411 & \underline{0.318} & 0.557 & 0.390 & 0.420 \\
GPT-4o~\cite{singh2025openai}        & 0.352 & 0.386 & 0.338 & 0.300 & 0.291 & 0.249 & 0.527 & 0.418 & 0.359 \\
MiniMax-M2.7~\cite{chen2026minimax}  & 0.257 & 0.243 & 0.246 & 0.186 & 0.181 & 0.179 & 0.402 & 0.291 & 0.249 \\
Kimi-K2.6~\cite{team2026kimi}     & 0.170 & 0.262 & 0.200 & 0.137 & 0.177 & 0.127 & 0.330 & 0.298 & 0.212 \\
Qwen-3.0-30B-A3B-Instruct~\cite{yang2025qwen3} & 0.411 & 0.405 & 0.395 & 0.354 & 0.306 & 0.243 & 0.617 & \underline{0.475} & 0.403 \\
Qwen-3.5-35B-A3B~\cite{yang2025qwen3} & 0.394 & 0.360 & 0.362 & 0.319 & 0.291 & 0.225 & 0.572 & 0.426 & 0.372 \\
Qwen-3.5-35B-A3B-Thinking~\cite{yang2025qwen3} & 0.475 & 0.368 & \underline{0.416} & 0.350 & \underline{0.362} & 0.301 & 0.602 & 0.454 & 0.419 \\
\bottomrule
\end{tabular}
}
\caption{Main Results on ToolBench-X. Bold text indicates the best result in each category, while underlined text indicates the second-best result.}
\label{tab:full_accuracy}
\end{table*}

Figure \ref{statistic} presents an overview of the benchmark composition. The tasks span seven diverse real-world domains. Reliability hazards are represented across all five uncertainty types, with cross-source, execution, and output uncertainty comprising the largest categories, while invocation and specification uncertainty provide complementary coverage. The benchmark further incorporates a balanced mix of sequential, parallel, and hybrid workflow structures. Most tasks involve multi-step tool usage, predominantly concentrated around four-tool workflows, with additional variability in the length of tool chains.

\section{Experiments}

\subsection{Experimental Setup}
We conduct a comprehensive evaluation on ToolBench-X by benchmarking twelve prominent large language models, covering both proprietary and open-source systems. The complete list of evaluated models is provided in Table~\ref{app_tab:models}. We report final-task accuracy as the primary metric. A task is considered correct if the execution state recorded by the backend at task completion matches the ground truth, or if the model's final response explicitly contains the ground-truth answer.


\subsection{Main Result}

Table~\ref{tab:full_accuracy} reports the performance of twelve representative LLMs on ToolBench-X. The results show that robust tool use under exceptions remains highly challenging. Even the best-performing model, Doubao-Seed-2.0-Lite, solves only slightly more than half of the cases. Other frontier models also remain below the 0.50 accuracy threshold, with GPT-5.4 achieving 0.453, followed by DeepSeek-V4-Pro, GLM-5.1, Gemini-3.1-Flash, and Claude-Sonnet-4.6. This suggests that no proprietary model family has yet addressed exception-aware tool use in unreliable tool-use environments.

Open-source models substantially narrow this gap. Qwen-3.5-35B-A3B-Thinking reaches an accuracy of 0.419, outperforming GPT-4o and approaching several closed-source systems. Comparisons within the Qwen family further suggest that explicit reasoning is more effective than parameter scaling alone: the thinking-enabled 35B model improves over its non-thinking counterpart by 4.7 percentage points, while the non-thinking 35B model underperforms the smaller Qwen-3.0-30B-A3B-Instruct.

Task-level results reveal that Parallel tasks are the easiest, with an average accuracy of 0.421, followed by Mixture and Sequential tasks. The performance gap reaches as much as fifteen percentage points for stronger models, confirming that errors accumulate when tool calls depend on previous outputs. Exception-level results show even sharper disparities. Output exceptions are handled relatively well, with an average accuracy of 0.581, whereas Specification exceptions are only moderately addressed. Execution, Cross-Source, and especially Invocation exceptions remain the dominant failure modes. The gap of more than thirty percentage points between Output and Invocation exceptions suggests that current LLMs are considerably better at post-hoc interpretation of tool results than at proactively generating correct tool calls under unreliable environments.

Overall, model rankings are relatively stable across task types but fluctuate markedly across exception types, while no model achieves an overall score above 0.60. Moreover, large-scale closed-source models do not exhibit a clear advantage over open-source counterparts. These results indicate that robust task execution in untrusted tool environments is a capability that must be explicitly targeted, rather than an ability that can be obtained through parameter scaling alone.

\subsection{Further Analysis}

\begin{figure}[t] 
    \centering
    \includegraphics[width=0.46\textwidth, height=0.2\textheight]{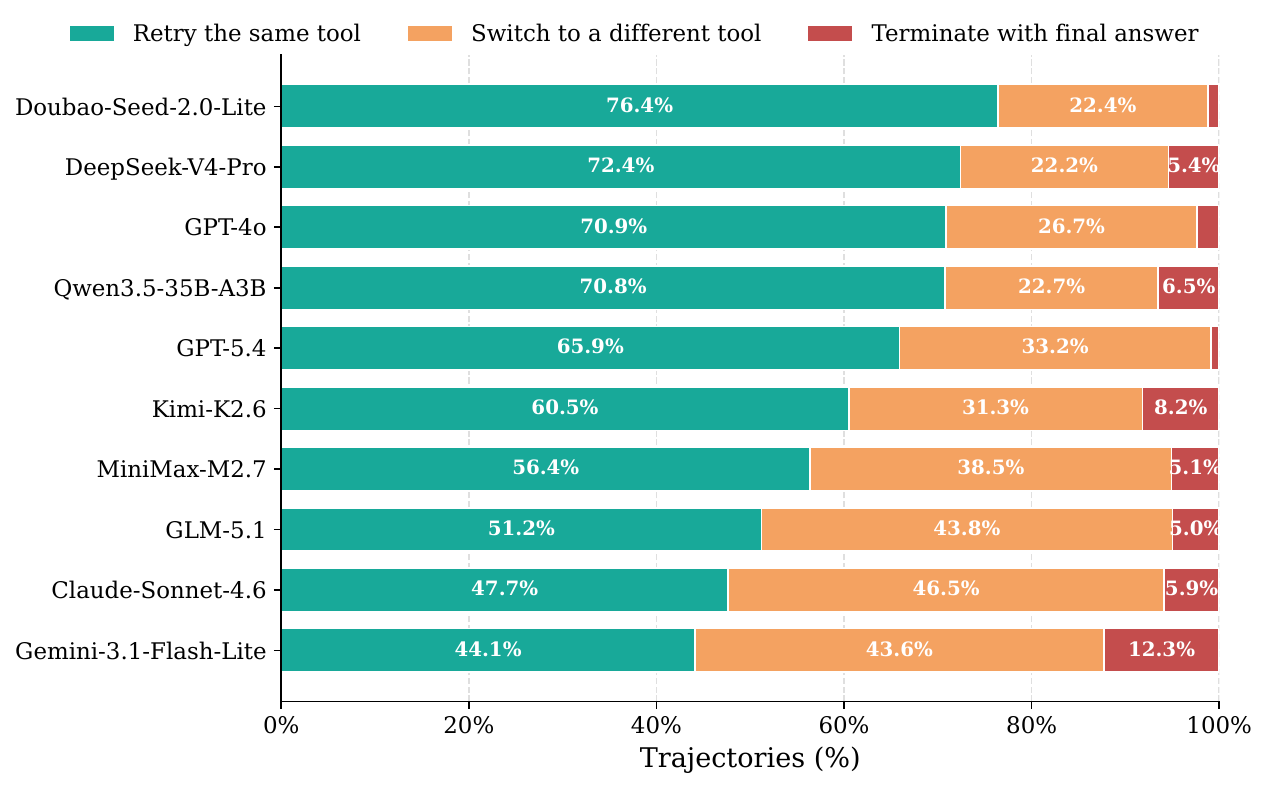} 
    \caption{Post-error behaviors after the first failed tool response.}
    \label{figure3}
\end{figure}
\textbf{Post-Failure Recovery Behavior.}
Aggregate accuracy alone does not explain why agents fail. We therefore examine the action immediately following the first failed tool call. Figure~\ref{figure3} shows that retrying the same tool dominates across all models, accounting for 44\% to 76\% of post-failure trajectories, while direct termination remains below 12\%. However, this behavior pattern is only weakly associated with overall accuracy in Table~\ref{tab:full_accuracy}. High retry rates appear in both the strongest model, Doubao-Seed-2.0-Lite with accuracy 0.513, and weaker models such as GPT-4o with accuracy 0.359. Similarly, switch-heavy models such as Claude-Sonnet-4.6 and Gemini-3.1-Flash achieve only mid-level accuracy. These results indicate that recovery quality depends not on retrying or switching, but on whether the action matches the underlying hazard. Failures mainly reflect inaccurate hazard diagnosis, demonstrating that diagnosis is the central bottleneck.

\label{sec:no_hint_with_hint}

\begin{figure}[t] 
    \centering
    \includegraphics[width=0.46\textwidth, height=0.18\textheight]{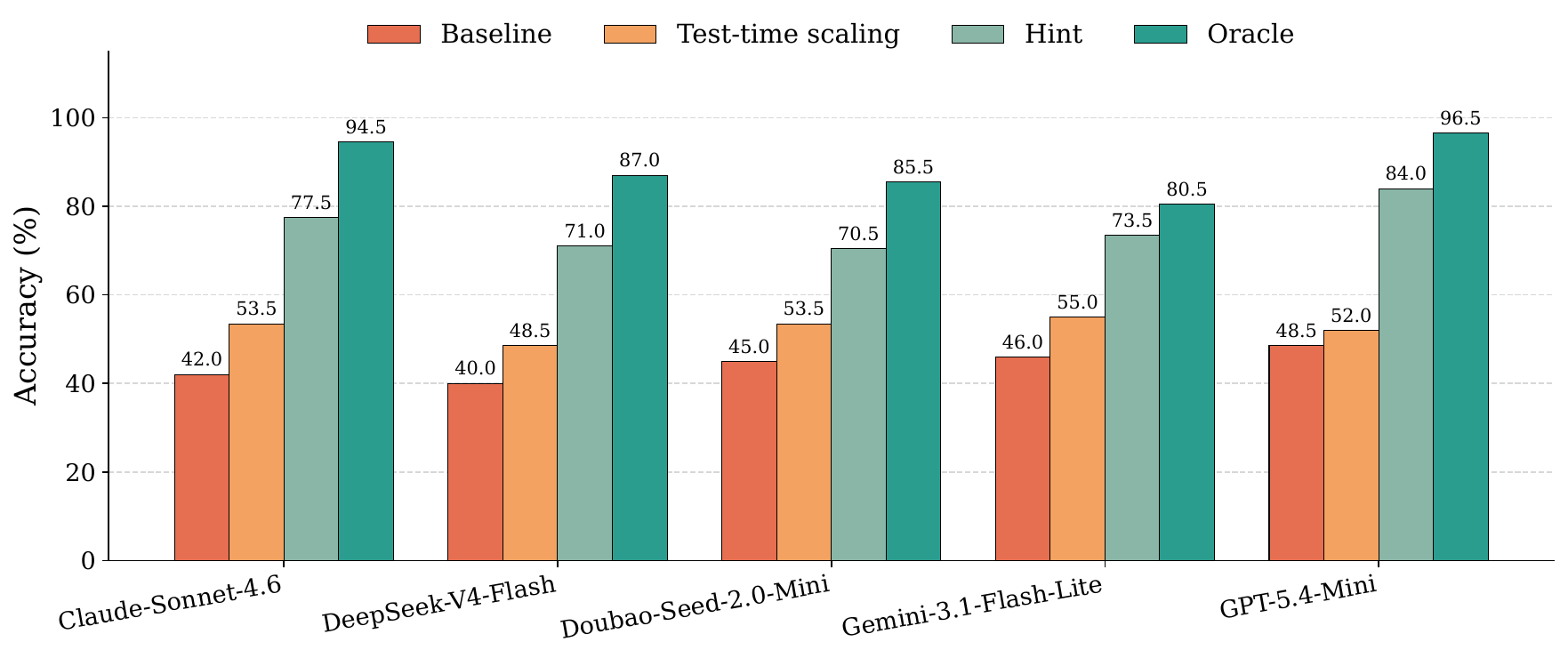} 
    \caption{Overall accuracy on the 200-task subset across five models.
Baseline is the exception-injected setting, Oracle is the clean upper
bound, and Test-time scaling and Hint are recovery strategies.}
    \label{figure4}
\end{figure}

\paragraph{Are failures unrecoverable, or merely undiagnosed?}
Agents fail substantially under tool exceptions, but aggregate accuracy alone cannot explain why. Low scores may reflect cases that become intrinsically unsolvable under a corrupted tool environment, or cases that remain solvable but require the agent to diagnose the tool environment hazard and adapt its strategy. To obtain credible conclusions while keeping evaluation cost manageable, we uniformly sample a \textbf{subset of 200 tasks} from the full benchmark as our evaluation pool, preserving the original distribution of task types and exception categories. The standard setting with injected exceptions on this subset serves as the \textbf{Baseline}. We further introduce three diagnostic settings. In \textbf{Hint}, after the agent fails a task, the model is informed of the current problem in the tool environment. In \textbf{Test Time Scaling} (TTS), after the agent fails a task, the model is given an additional budget of 10 interaction rounds to retry without any hint. In \textbf{Oracle}, the same tasks are evaluated in the original clean tool environment without exception injection, providing an upper bound on achievable performance. The gap between Baseline and Oracle measures the performance loss caused by tool exceptions, while the relative positions of Hint and TTS within this gap help distinguish failures caused by poor diagnosis from those caused by insufficient compute.

\paragraph{Diagnosis is the dominant bottleneck.}
Figure~\ref{figure4} reveals a sizable gap of roughly 35 to 50 points between Baseline and Oracle, confirming that exception injection is a substantive source of failure rather than a minor perturbation. Hint close the bulk of this gap. Hint lifts Baseline accuracy by 25.5 to 35.5 absolute points, recovering roughly 60 to 80 percent of the lost accuracy across all five models. The improvement is consistent across capability tiers, from DeepSeek-V4-Flash to GPT-5.4-Mini, indicating that diagnosis is a uniformly limiting factor rather than a weakness specific to smaller models. Although each hint provides only a brief explanation of the anomaly in the tool environment, this signal alone is sufficient to turn most failures into successes. Most no-hint failures are therefore not on irrecoverable tasks but on tasks where the agent cannot identify the hazard on its own. Nevertheless, a residual gap between Hint and Oracle remains, suggesting that certain hazards require adaptive recovery behaviors that current models still fail to execute reliably.

\paragraph{Compute alone does not explain the Hint gain.}
TTS isolates the effect of additional compute by granting extra inference rounds while withholding any information about anomalies in the tool environment. Thus, any remaining advantage of Hint over TTS can be attributed to diagnostic information rather than to additional reasoning steps. The results indicate that compute alone is a weak intervention: TTS improves Baseline accuracy by only 3.5–11.5 percentage points, while Hint consistently outperforms TTS by 24–32 points. This gap persists even among the strongest models in our evaluation. GPT-5.4-Mini achieves 0.840 accuracy with Hint but only 0.520 with TTS, while Claude-Sonnet-4.6 reaches 0.775 versus 0.535. Taken together with the strong Hint performance, these findings identify diagnosis, rather than recovery, as the primary bottleneck. Once an agent knows which hazard it faces, it can often recover effectively. Robust exception handling therefore requires hazard-aware sensing and self-diagnosis, rather than simply scaling the inference budget.

\begin{table}[t]
\centering

\resizebox{\columnwidth}{!}{%
\begin{tabular}{lccc c}
\toprule
\multirow{2}{*}{\textbf{MLLMs}} 
& \multicolumn{3}{c}{\textbf{Tools per Task}} 
& \multirow{2}{*}{\textbf{Overall}} \\
\cmidrule(lr){2-4}
& $\leq 3$ & $4$ & $\geq 5$ & \\
\midrule
DeepSeek-V4-Pro & 0.413 & 0.432 & 0.399 & 0.425 \\
GPT-5.4 & 0.442 & 0.461 & 0.422 & 0.453 \\
Doubao-Seed-2.0-Lite & 0.577 & 0.515 & 0.462 & 0.513 \\
Claude-Sonnet-4.6 & 0.442 & 0.414 & 0.376 & 0.410 \\
Gemini-3.1-Flash-Lite & 0.442 & 0.433 & 0.318 & 0.416 \\
GLM-5.1 & 0.490 & 0.428 & 0.335 & 0.420 \\
GPT-4o & 0.423 & 0.363 & 0.301 & 0.359 \\
MiniMax-M2.7 & 0.269 & 0.251 & 0.225 & 0.249 \\
Kimi-K2.6 & 0.269 & 0.203 & 0.220 & 0.212 \\
Qwen-3.0-30B-A3B-Instruct & 0.404 & 0.415 & 0.347 & 0.403 \\
Qwen-3.5-35B-A3B & 0.404 & 0.370 & 0.358 & 0.372 \\
Qwen-3.5-35B-A3B-Thinking & 0.471 & 0.425 & 0.358 & 0.419 \\
\midrule
\textit{N} & 104 & 829 & 173 & 1106 \\
\bottomrule
\end{tabular}%
}
\caption{Overall accuracy on the full benchmark by tool-count complexity. Tasks are grouped into low-complexity ($\leq 3$ tools), medium-complexity (4 tools), and high-complexity ($\geq 5$ tools) buckets.}
\label{tab:tool_count_scaling}
\end{table}

\paragraph{Do longer tool chains compound exception hazards?}

A natural concern is that the failures we observe are amplified by task length: each additional tool call introduces another point of potential exception, another intermediate observation that must be checked, and another opportunity for errors to propagate downstream. We therefore examine how reliability scales with the number of tools required by a task. Because exact tool-count bins are unevenly populated, we group tasks into three buckets — \textbf{low} ($\leq 3$ tools), \textbf{medium} ($4$ tools), and \textbf{high} ($\geq 5$ tools), which preserves statistical power in each bucket while exposing the trend across complexity levels.

Table~\ref{tab:tool_count_scaling} reports accuracy under this grouping. Most models show a clear decline from low- to high-complexity tasks. For example, Doubao-Seed-2.0-Lite drops from $0.577$ to $0.462$, Gemini-3.1-Flash-Lite from $0.442$ to $0.318$, GLM-5.1 from $0.490$ to $0.335$, and GPT-4o from $0.423$ to $0.301$. These declines indicate that exception handling becomes harder as agents must coordinate longer chains of tool calls.

The trend is not perfectly monotonic across all models, suggesting that tool count is not the only source of difficulty. Exception type, workflow structure, and dependencies between calls also affect robustness. Nevertheless, the overall pattern supports a compounding-risk interpretation: as tool chains grow longer, a single mishandled exception is more likely to corrupt subsequent reasoning. Reliable agent behavior under tool uncertainty therefore requires robust per-call diagnosis and recovery, not merely stronger final-answer generation.


\begin{figure}[t]
    \centering
    \includegraphics[width=0.98\linewidth, height=0.25\textheight]{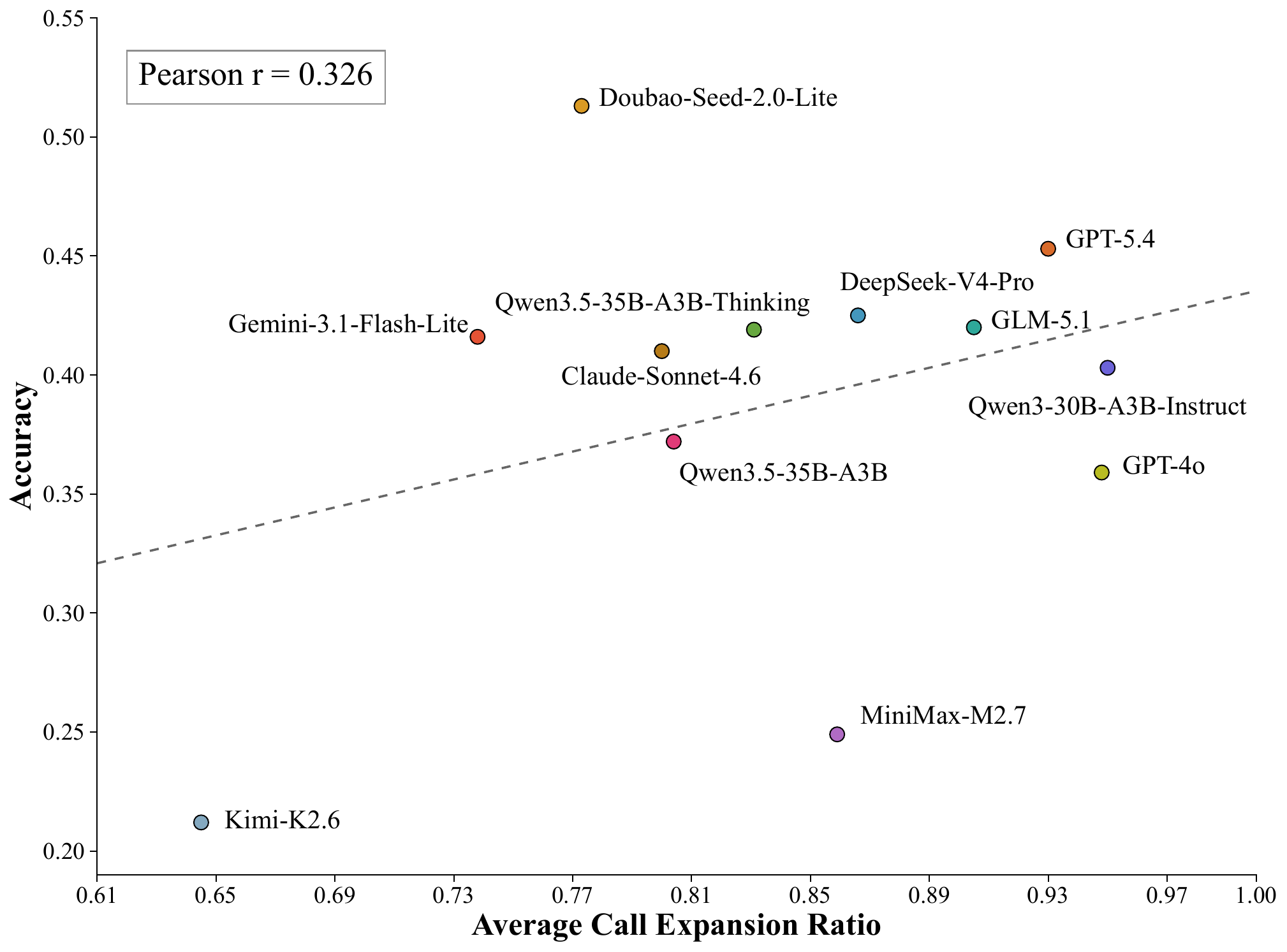}
    \caption{
    Process efficiency versus accuracy on the full benchmark. 
    }
    \label{fig:efficiency_accuracy_scatter}
\end{figure}

\paragraph{Does robustness come from \emph{more} calls, or \emph{better} calls?}
\label{sec:efficiency_accuracy}
A natural hypothesis is that the more accurate models in Table~\ref{tab:full_accuracy} simply call tools more aggressively. To test this, we measure each model's tool-invocation intensity using the Average Call Expansion Ratio:
\[
\overline{R} \;=\; \frac{1}{|\mathcal{T}|}\sum_{t \in \mathcal{T}} \frac{C_t}{K_t},
\]
where $C_t$ denotes the number of tool calls made on task $t$, and $K_t$ denotes the number of tools specified by that task. We use $\overline{R}$ as a process-level diagnostic rather than a capability score.

Figure~\ref{fig:efficiency_accuracy_scatter} compares accuracy with $\overline{R}$ across models. The association is weak, with a descriptive model-level Pearson $r=0.326$ computed from each model's average performance over 1,106 tasks. The ordering of several models further contradicts a volume-based explanation. Doubao-Seed-2.0-Lite achieves the highest accuracy of $0.513$ with a relatively low expansion ratio of $0.773$, whereas GPT-4o and Qwen-3.0-30B-A3B-Instruct invoke tools more frequently, with expansion ratios of $0.948$ and $0.950$, yet obtain lower accuracy. These results indicate that robustness is not primarily explained by invocation volume. Effective agents recover by using additional observations to revise their reasoning, rather than by merely increasing the number of tool calls.

\subsection{Error Analysis}
\label{sec:error_analysis}

\begin{figure}[t]
    \centering
    \includegraphics[width=\linewidth, height=0.22\textheight]{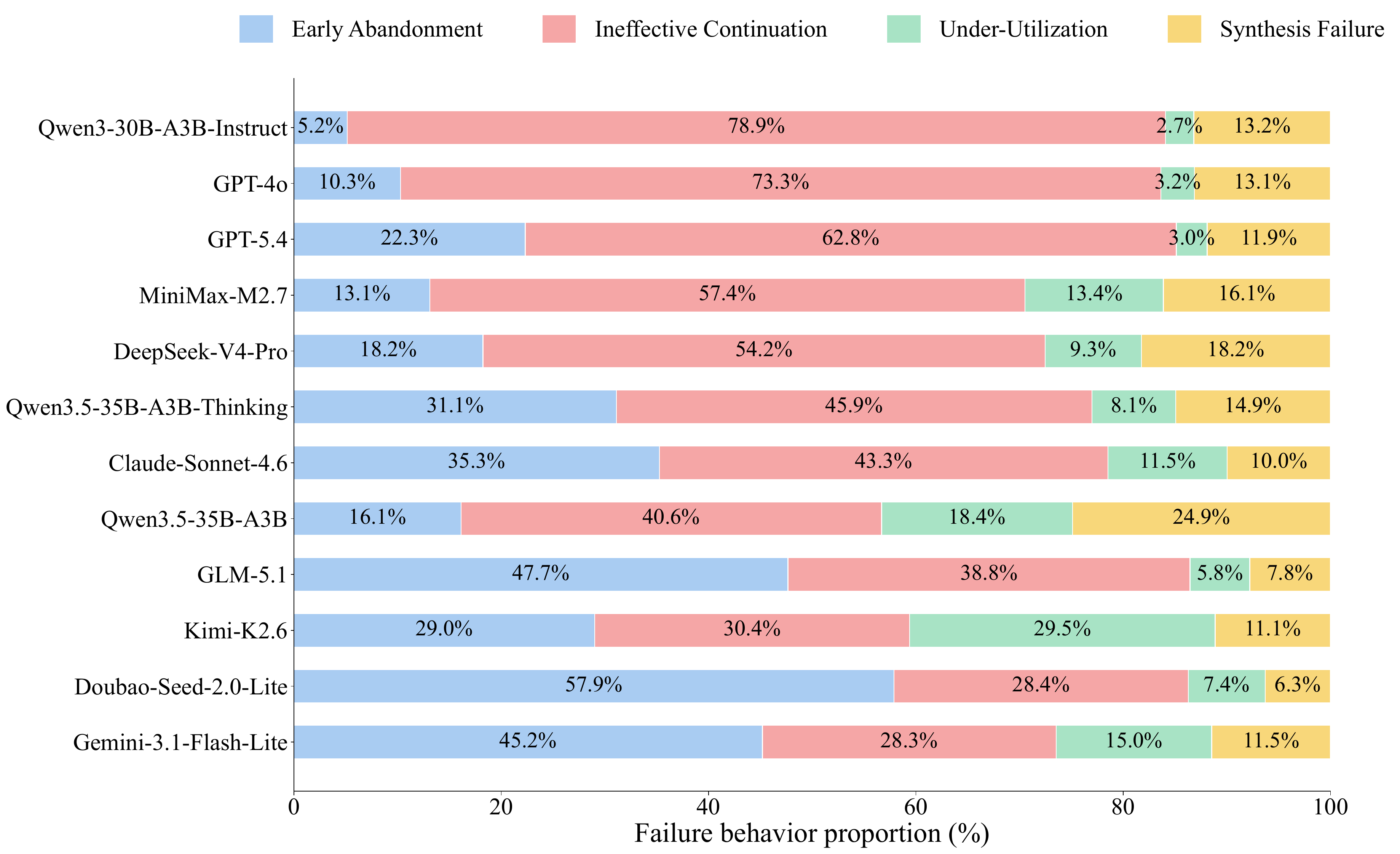}
    \caption{
    Failure behavior taxonomy by model. 
    }
    \label{fig:failure_taxonomy_stacked}
\end{figure}

Final-answer accuracy indicates whether an agent succeeds, but not how failure unfolds. To diagnose failed no-hint trajectories, we use four observable signals: final-answer validity, whether tool use continues after the first error, the number of post-error calls $A_t$, and the call expansion ratio $R_t$. With median thresholds of $A_t=3$ and $R_t=1.00$, we group failures into four behaviors: \textbf{Ineffective Continuation}, where agents keep using tools intensively but fail to recover; \textbf{Early Abandonment}, where agents stop early or return invalid answers; \textbf{Under-utilization}, where tool coverage remains insufficient; and \textbf{Answer Synthesis Failure}, where agents use tools but fail to produce the correct final answer.

Figure~\ref{fig:failure_taxonomy_stacked} shows that \textbf{Ineffective Continuation} is the dominant failure mode, accounting for $48.6\%$ of failed trajectories, followed by \textbf{Early Abandonment} at $26.7\%$. This indicates that many failures arise not from immediate stopping, but from unproductive recovery attempts after tool errors. The model-level breakdown further reveals that similar final accuracies can hide distinct failure mechanisms. Qwen-3.0-30B-A3B-Instruct and GPT-4o show large shares of \textbf{Ineffective Continuation}, suggesting repeated but poorly adapted retries, whereas Kimi-K2.6 shows more \textbf{Under-utilization}, indicating insufficient tool exploration. These patterns motivate trajectory-level diagnostics beyond aggregate accuracy.

\section{Conclusion}

This work introduces ToolBench-X, a benchmark for evaluating tool-using agents under recoverable tool-environment unreliability. By incorporating structured hazards into otherwise solvable tasks, ToolBench-X shifts evaluation from task completion with reliable tools to agents’ ability to diagnose unreliable tool behavior, adapt their strategies, recover from failures, and ultimately complete the task. Experiments show that current agents remain fragile, with failures driven more by limited anomaly awareness and ineffective recovery than by tool-use volume or inference budget. These findings suggest that robust tool use requires reliability-aware agent design, including better uncertainty estimation, verification, fallback, and recovery mechanisms. ToolBench-X provides a foundation for systematically evaluating and developing trustworthy, resilient tool-using agents in real-world settings.

\bibliography{aaai2027}

\appendix

\section{Human Validation and Quality Control}
We validate each generated benchmark instance through a multi-stage human review process. The review team consists of five reviewers, each of whom is either a Ph.D. student or a Ph.D. degree holder. The review spans 20 working days, with an 8-hour schedule per day, for a total of 800 person-hours. Each reviewer is paid at least the applicable local minimum hourly wage.

\paragraph{Review process.} Each candidate instance is assigned to a primary reviewer, who examines the task specification, tool implementation, injected reliability exception, reference answer, and recovery path. If an instance passes the initial review but contains ambiguous instructions, non-trivial recovery behavior, cross-source evidence, or possible answer leakage, it is sent to a second reviewer for independent validation. Disagreements are resolved through discussion by at least two reviewers. Instances without consensus are removed from the benchmark.

\paragraph{Review criteria.} For each candidate instance, reviewers verify that the user task is clear and has a unique expected answer, that all associated Python tools run deterministically in a clean environment, that the reference answer is consistent with the tool path without exception injection, and that the injected exception matches its intended category, including specification drift, invocation error, execution failure, output drift, or cross-source conflict. These five hazards mark different points at which a task can fail, so we do not apply every hazard type to every task. We assign a hazard only when it fits the task structure and when the original reference answer can still be reached through at least one valid recovery path. Reviewers also verify that the injected instance retains at least one valid recovery path, that the required recovery behavior can be achieved through retry, fallback, normalization, validation, or cross-checking, that the prompt, tool names, tool descriptions, hints, and metadata do not directly reveal the final answer, and that exact-match evaluation agrees with human semantic judgment.

\paragraph{Automatic recovery-path validation.} For each candidate instance, we automatically execute both the clean path without exception injection and the expected recovery path in the injected tool environment. An instance passes this check only if the clean path produces the reference answer and the injected environment still contains a valid recovery path to the same answer.

\paragraph{Answer-leakage check.} We use both automatic and human checks for answer leakage. The automatic checker searches the user prompt, tool descriptions, tool metadata, exception messages, and recovery hints for exact and normalized forms of the reference answer. Normalization includes lowercase conversion, whitespace removal, punctuation removal, unit stripping, and common numeric format conversion. Human reviewers further check whether the final answer can be inferred without executing the tools. Instances with direct leakage are revised or removed.

\paragraph{Candidate-instance statistics.} The initial generation stage produces 2,610 raw task items. We first apply preliminary screening for task validity, duplication, tool executability, and deterministic clean-path execution, yielding 1,250 candidate instances across 84 subtopics and three workflow types. At most five candidates are retained for each combination of task type and subtopic. These candidates then undergo human review and automatic validation. Among the 1,250 candidates, 1,142 pass the automatic recovery-path check, and 1,196 pass the answer-leakage check. Candidates that fail either check are manually inspected and are revised, regenerated, or removed depending on whether the issue can be fixed without changing the task intent. After this process, the final benchmark contains 1,106 instances, including 378 sequential, 358 parallel, and 370 mixture tasks. The overall retention rate is 88.5\% over the candidate pool. According to the review logs, 996 instances are retained without substantive changes (79.7\%), 72 are retained after revision (5.8\%), 38 are retained after regeneration (3.0\%), and 144 are removed (11.5\%).

\paragraph{Double review and agreement.} A total of 200 instances are independently reviewed by two reviewers, covering 18.1\% of the retained benchmark. Before discussion, reviewer agreement is measured based on accept, revise, or remove decisions and the assigned exception categories. The initial agreement rate is 91.5\%. After discussion and correction, all retained instances receive consensus approval.

\paragraph{Agreement between exact match and semantic judgment.} To test whether exact-match evaluation rejects semantically correct answers, we draw a stratified random sample of 100 retained instances and compare the backend exact-match results with independent human semantic judgments. The two criteria agree in 97 cases, giving an agreement rate of 97.0\%. Disagreements are manually inspected. If an exact-match target is too brittle or allows multiple valid surface forms, we revise the reference-answer format and recheck the instance.

\section{Additional Dataset Statistics Details} 
Tables~\ref{tab:task_topic_subtopic_statistics} and~\ref{tab:task_topic_subtopic_statistics_continued} report the hierarchical composition of the final dataset by task type, main topic, and subtopic. The dataset contains 1,106 retained tasks, including 378 sequential, 358 parallel, and 370 mixture tasks. Overall, it covers 7 main topics and 84 unique subtopics.

Each table entry gives the number of retained tasks for a particular task-type--subtopic combination. Although up to five tasks were initially generated for each combination, the final counts vary because tasks that did not satisfy the benchmark validation and executability requirements were excluded. Across all task types, the largest topic groups are Commerce \& Transactions (234 tasks), Entertainment \& Media (215), and Health \& Wellness (192).

\section{Detailed Experimental Settings}
\label{detailed_experimental_settings}
We conduct a comprehensive evaluation on seven
commercial models and seven representative open-source large language models. The complete list of models evaluated is provided in Table~\ref{app_tab:models}.

All models were accessed via a unified OpenAI‑compatible API and used their default temperature. We allowed 10 rounds per task, 8 parallel workers, a maximum of 8,192 output tokens, and a 120‑second timeout per request. To preserve diagnostic granularity and recovery observability, at most one tool call, retry, fallback, or finish was permitted per round. Each task required a single definite output (e.g., amount, date, count); accordingly, we adopted a strict exact‑match criterion: after stripping whitespace, the model output must be identical to the answer string. API requests were retried up to three times with exponential backoff (2–10 seconds); persistent failures were recorded as task failures.

\section{Case Studies}
We present six representative no-hint trajectory pairs from the matched 200-task subset in Fig.~\ref{fig:case_study}. Each pair uses the same task and injected tool implementation, comparing a model that reaches the canonical benchmark answer with one that fails under the same hazard. Hazard labels are assigned according to the semantic location of the failure and the recovery path required to reach the answer, rather than the surface Python exception alone. Thus, the same exception type may require different recovery behaviors across cases.

\section{Prompt Templates}
\subsection{Task Generation Prompts}
Table~\ref{tab:seq_task_generation_prompt} to \ref{tab:mix_task_generation_prompt} show the prompt for task generation.
\subsection{Tool Generation Prompts}
Table~\ref{tab:tool_generation_prompt} shows the prompt for tool generation.

\subsection{Hazard Injection and Hint Generation Prompts.}
Table~\ref{tab:hazard_injection_prompt} shows the prompt for reliability hazard injection and hint generation.

\subsection{Policy Decision Prompts}
Table~\ref{tab:evaluation_policy_prompt} shows the prompt for policy decision.

\subsection{Tool Arguments Prompts}
Table~\ref{tab:tool_argument_prompt} shows the prompt for policy decision.

\subsection{Final-Answer Fallback Prompts}
Table~\ref{tab:final_answer_prompt} shows the prompt for final-answer fallback.

\subsection{Test-Time Scaling Prompts}
Table~\ref{tab:test_time_scaling_prompt} shows the prompt for test-time scaling.

\input{Tables/add_main_subtopic}
\input{Tables/Faliure_behavior}
\input{Tables/models}

\input{Figures/case_study}

\input{Tables/prompt_seq_task}
\input{Tables/prompt_par_task}
\input{Tables/prompt_mix_task}
\input{Tables/prompt_tool_generation}
\input{Tables/prompt_hazard_injection}
\input{Tables/prompt_policy}
\input{Tables/prompt_tool_argument}
\input{Tables/prompt_final_answer}
\input{Tables/prompt_test_time_scaling}



\end{document}

%% file: Tables/add_main_subtopic.tex

\begin{table}[htbp]
\centering
\caption{Number of retained tasks for each task-type, main-topic, and subtopic combination. S, P, and M denote sequential, parallel, and mixture tasks.}
\label{tab:task_topic_subtopic_statistics}
\scriptsize
\setlength{\tabcolsep}{2pt}
\renewcommand{\arraystretch}{0.78}

\resizebox{\columnwidth}{!}{%
\begin{tabular}{@{}lrrrr@{}}
\hline
\textbf{Main Topic / Subtopic}
& \textbf{S} & \textbf{P} & \textbf{M} & \textbf{Total} \\
\hline

\multicolumn{5}{@{}l}{\textbf{Commerce \& Transactions}} \\
Cart and checkout issue resolution      & 3 & 5 & 5 & 13 \\
Cart and checkout issue support         & 5 & 5 & 5 & 15 \\
Coupon and discount lookup              & 5 & 5 & 5 & 15 \\
Gift purchase coordination              & 4 & 4 & 3 & 11 \\
Invoice and receipt organization        & 5 & 4 & 4 & 13 \\
Invoice and receipt retrieval           & 5 & 5 & 5 & 15 \\
Online order tracking                   & 5 & 5 & 5 & 15 \\
Price drop monitoring                   & 5 & 1 & 5 & 11 \\
Product search and comparison           & 5 & 5 & 4 & 14 \\
Return and refund handling              & 5 & 4 & 5 & 14 \\
Return and refund management            & 5 & 4 & 5 & 14 \\
Service booking and payment follow-up   & 5 & 4 & 5 & 14 \\
Shipping status tracking                & 5 & 5 & 5 & 15 \\
Subscription plan review                & 5 & 5 & 5 & 15 \\
Subscription purchase decisions         & 4 & 5 & 5 & 14 \\
Vendor and marketplace comparison       & 5 & 4 & 5 & 14 \\
Vendor and marketplace selection        & 5 & 3 & 4 & 12 \\
\textit{Subtotal}                        & \textbf{81} & \textbf{73} &
\textbf{80} & \textbf{234} \\
\hline

\multicolumn{5}{@{}l}{\textbf{Data \& Analytics}} \\
Customer feedback categorization         & 5 & 4 & 5 & 14 \\
Dashboard metric tracking                & 5 & 4 & 5 & 14 \\
Data entry validation and error checking & 5 & 5 & 5 & 15 \\
Inventory data monitoring                & 4 & 5 & 4 & 13 \\
KPI comparison across time periods       & 4 & 4 & 4 & 12 \\
Report generation and scheduling         & 5 & 5 & 5 & 15 \\
Sales data summarization                 & 5 & 5 & 5 & 15 \\
Spreadsheet cleanup and formatting       & 4 & 4 & 4 & 12 \\
Survey response analysis                 & 5 & 4 & 5 & 14 \\
Trend identification in business data    & 4 & 5 & 5 & 14 \\
\textit{Subtotal}                         & \textbf{46} & \textbf{45} &
\textbf{47} & \textbf{138} \\
\hline

\multicolumn{5}{@{}l}{\textbf{Entertainment \& Media}} \\
Book and reading list management        & 3 & 4 & 4 & 11 \\
Content release date tracking           & 5 & 2 & 5 & 12 \\
Event and show discovery                & 4 & 5 & 5 & 14 \\
Event and ticket planning               & 5 & 5 & 3 & 13 \\
Family-friendly content filtering       & 5 & 5 & 3 & 13 \\
Game selection by mood or group size    & 5 & 4 & 5 & 14 \\
Gaming session planning                 & 5 & 5 & 4 & 14 \\
Media release date tracking             & 5 & 4 & 5 & 14 \\
Media subscription comparison           & 5 & 5 & 5 & 15 \\
Movie and TV recommendation planning    & 5 & 4 & 5 & 14 \\
Music playlist organization             & 5 & 5 & 5 & 15 \\
Photo and video library organization    & 5 & 5 & 4 & 14 \\
Podcast discovery by interest           & 5 & 5 & 5 & 15 \\
Streaming subscription management       & 5 & 3 & 5 & 13 \\
Streaming watchlist management          & 4 & 2 & 5 & 11 \\
Watchlist and reading list organization & 3 & 5 & 5 & 13 \\
\textit{Subtotal}                        & \textbf{74} & \textbf{68} &
\textbf{73} & \textbf{215} \\
\hline

\end{tabular}%
}

\end{table}


\begin{table}[H]
\centering
\caption{Number of retained tasks by task type, main topic, and subtopic (continued).}
\label{tab:task_topic_subtopic_statistics_continued}
\scriptsize
\setlength{\tabcolsep}{2pt}
\renewcommand{\arraystretch}{0.78}

\resizebox{\columnwidth}{!}{%
\begin{tabular}{@{}lrrrr@{}}
\hline
\textbf{Main Topic / Subtopic}
& \textbf{S} & \textbf{P} & \textbf{M} & \textbf{Total} \\
\hline

\multicolumn{5}{@{}l}{\textbf{Finance \& Economics}} \\
Bank transaction monitoring          & 5 & 4 & 5 & 14 \\
Basic investment portfolio tracking  & 4 & 2 & 5 & 11 \\
Bill payment planning                & 4 & 5 & 3 & 12 \\
Credit card usage monitoring         & 5 & 4 & 5 & 14 \\
Currency conversion for spending     & 4 & 5 & 3 & 12 \\
Expense categorization and review    & 1 & 5 & 5 & 11 \\
Household cost comparison            & 3 & 5 & 3 & 11 \\
Loan payment estimation              & 3 & 4 & 3 & 10 \\
Personal budget tracking             & 4 & 5 & 5 & 14 \\
Savings goal planning                & 5 & 4 & 4 & 13 \\
Savings goal tracking                & 2 & 2 & 4 & 8 \\
Tax document organization            & 4 & 2 & 3 & 9 \\
\textit{Subtotal}                    & \textbf{44} & \textbf{47} &
\textbf{48} & \textbf{139} \\
\hline

\multicolumn{5}{@{}l}{\textbf{Government \& Public Services}} \\
Court and civic appointment reminders       & 5 & 5 & 4 & 14 \\
Emergency service and alert information     & 5 & 3 & 4 & 12 \\
License and ID renewal tracking             & 5 & 5 & 5 & 15 \\
Local government service requests           & 4 & 5 & 5 & 14 \\
Permit and license application tracking     & 5 & 4 & 5 & 14 \\
Public benefit application support          & 2 & 5 & 5 & 12 \\
Public health advisory lookup               & 5 & 4 & 5 & 14 \\
Public records request preparation          & 5 & 5 & 3 & 13 \\
Public transit service updates              & 5 & 5 & 4 & 14 \\
School district and enrollment information  & 5 & 1 & 4 & 10 \\
Tax filing deadline reminders               & 5 & 5 & 5 & 15 \\
Utility service setup or transfer           & 3 & 2 & 5 & 10 \\
Voter registration and election reminders   & 5 & 5 & 5 & 15 \\
\textit{Subtotal}                           & \textbf{59} & \textbf{54} &
\textbf{59} & \textbf{172} \\
\hline

\multicolumn{5}{@{}l}{\textbf{Health \& Wellness}} \\
Appointment scheduling and follow-up      & 5 & 5 & 4 & 14 \\
Exercise routine planning                 & 5 & 5 & 5 & 15 \\
Fitness routine planning                  & 4 & 4 & 3 & 11 \\
Habit building for healthy routines       & 5 & 5 & 5 & 15 \\
Meal and nutrition tracking               & 4 & 4 & 3 & 11 \\
Medication schedule reminders             & 5 & 5 & 5 & 15 \\
Mental wellness check-ins                 & 5 & 5 & 3 & 13 \\
Nutrition and meal planning               & 3 & 5 & 4 & 12 \\
Preventive care reminders                 & 5 & 5 & 5 & 15 \\
Recovery and rehabilitation tracking      & 5 & 5 & 3 & 13 \\
Sleep habit monitoring                    & 5 & 5 & 5 & 15 \\
Stress management support                 & 5 & 5 & 5 & 15 \\
Symptom logging and monitoring            & 5 & 5 & 5 & 15 \\
Wellness goal check-ins                   & 5 & 4 & 4 & 13 \\
\textit{Subtotal}                         & \textbf{66} & \textbf{67} &
\textbf{59} & \textbf{192} \\
\hline

\multicolumn{5}{@{}l}{\textbf{Information \& Communication}} \\
Announcement and newsletter drafting & 4 & 2 & 4 & 10 \\
Calendar invite coordination         & 4 & 2 & 0 & 6 \\
\textit{Subtotal}                    & \textbf{8} & \textbf{4} &
\textbf{4} & \textbf{16} \\
\hline

\textbf{Overall Total}
& \textbf{378}
& \textbf{358}
& \textbf{370}
& \textbf{1,106} \\
\hline

\end{tabular}%
}

\end{table}

%% file: Tables/Faliure_behavior.tex
\begin{table}[htbp]
\centering
\caption{Exact counts of mutually exclusive heuristic failure-behavior
categories among failed no-hint trajectories. Each failed trajectory is
assigned to exactly one category. EA: early abandonment; IC: ineffective
continuation; UU: under-utilization; SF: synthesis failure.}
\label{tab:failure-behavior-counts}
\resizebox{\linewidth}{!}{%
\begin{tabular}{@{}lrrrrr@{}}
\toprule
Model
& Failed
& EA
& IC
& UU
& SF \\
\midrule
DeepSeek-V4-Pro          & 636 & 116 & 345 &  59 & 116 \\
GPT-5.4                  & 605 & 135 & 380 &  18 &  72 \\
Qwen3-30B-A3B-Instruct       & 660 &  34 & 521 &  18 &  87 \\
Qwen3.5-35B-A3B              & 695 & 112 & 282 & 128 & 173 \\
Qwen3.5-35B-A3B-Thinking     & 643 & 200 & 295 &  52 &  96 \\
Doubao-Seed-2.0-Lite     & 539 & 312 & 153 &  40 &  34 \\
Claude-Sonnet-4.6        & 652 & 230 & 282 &  75 &  65 \\
Gemini-3.1-Flash-Lite    & 646 & 292 & 183 &  97 &  74 \\
GLM-5.1                  & 642 & 306 & 249 &  37 &  50 \\
GPT-4o                   & 709 &  73 & 520 &  23 &  93 \\
MiniMax-M2.7             & 831 & 109 & 477 & 111 & 134 \\
Kimi-K2.6                & 872 & 253 & 265 & 257 &  97 \\
\midrule
\textbf{Total}           & \textbf{8,130} & \textbf{2,172}
& \textbf{3,952} & \textbf{915} & \textbf{1,091} \\
\bottomrule
\end{tabular}
}
\end{table}

%% file: Tables/models.tex
\begin{table}[htbp]
\centering
\resizebox{\linewidth}{!}{%
\begin{tabular}{l p{\linewidth}}   
\hline
\textbf{Model Type} & \textbf{Models} \\
\hline
\multirow{3}{*}{Proprietary LLMs} & GPT-5.4, GPT-5.4-Mini, GPT-4o \\
                                  & Gemini-3.1-Flash, Claude-Sonnet-4.6 \\ & Doubao-Seed-2.0-Lite, Doubao-Seed-2.0-Mini \\
\hline
\multirow{3}{*}{Open-source LLMs} & DeepSeek-V4-Pro, DeepSeek-V4-Flash \\ & GLM-5.1, MiniMax-M2.7, Kimi-K2.6  \\ & Qwen3.0-30B-A3B-Instruct, Qwen-3.5-35B-A3B \\
\hline
\end{tabular}%
}
\caption{Evaluated large language models.}
\label{app_tab:models}
\end{table}

%% file: Figures/case_study.tex
\begin{figure*}[htbp]
    \centering
    \captionsetup[subfigure]{font=small,skip=2pt}
    \captionsetup{font=small}

    \begin{subfigure}{0.8\textwidth}
        \centering
        \includegraphics[width=\linewidth]{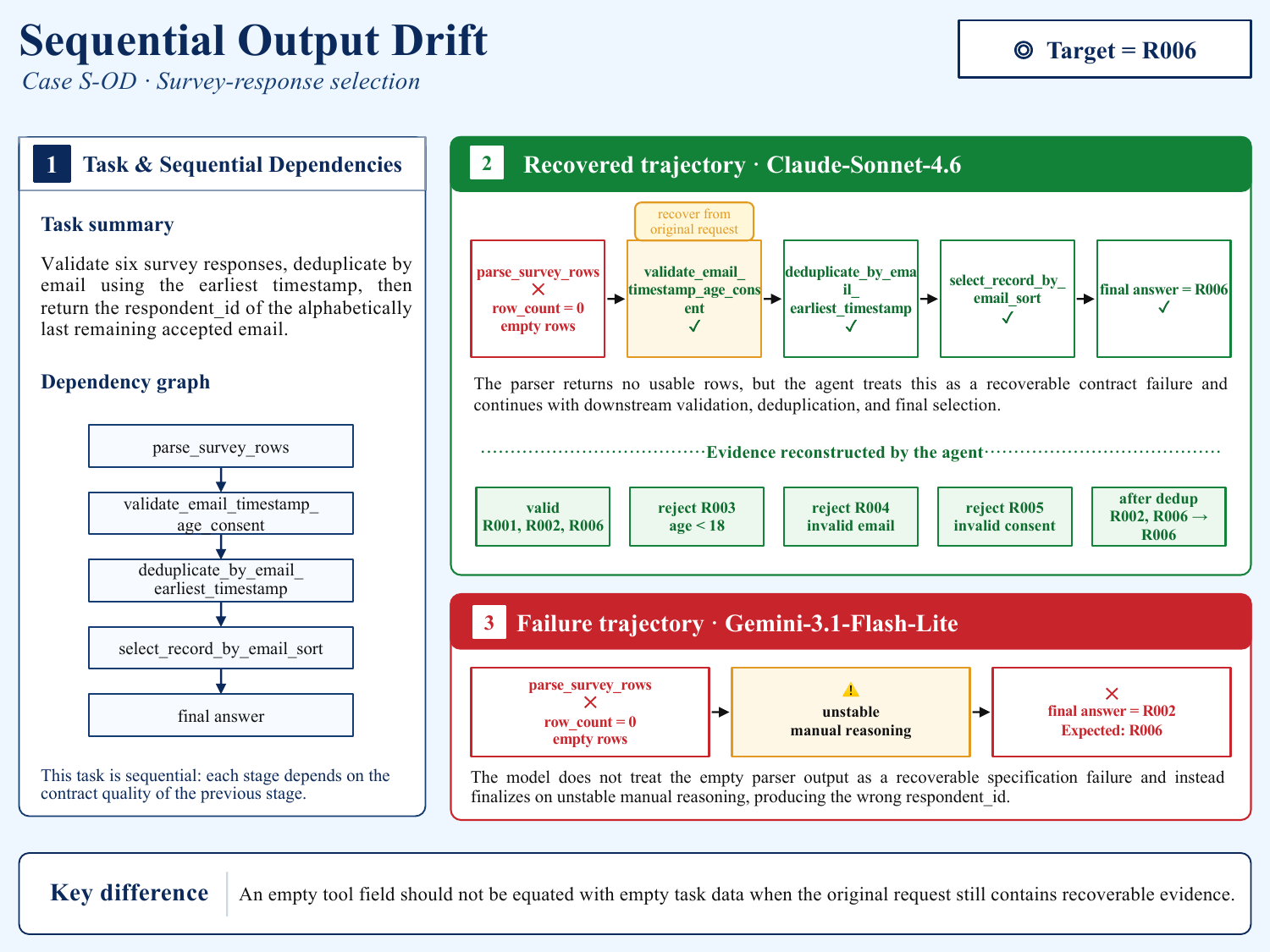}
        \caption{Sequential output drift}
        \label{fig:case1}
    \end{subfigure}

    \vspace{0.4em}

    \begin{subfigure}{0.8\textwidth}
        \centering
        \includegraphics[width=\linewidth]{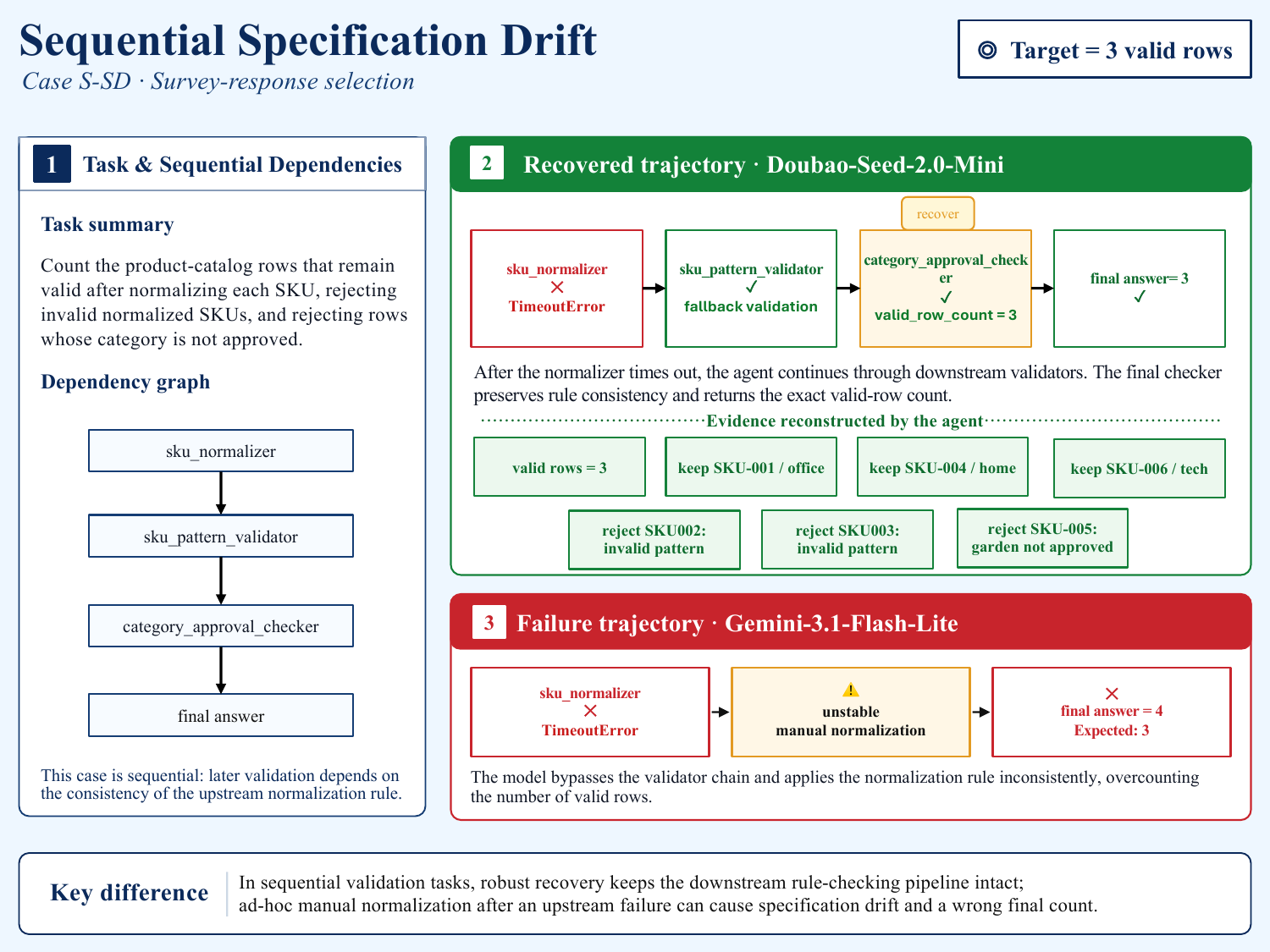}
        \caption{Sequential specification drift}
        \label{fig:case2}
    \end{subfigure}
    
\end{figure*}

\begin{figure*}[htbp]
    \ContinuedFloat
    \centering
    \captionsetup[subfigure]{font=small,skip=2pt}
    \captionsetup{font=small}

    \begin{subfigure}{0.8\textwidth}
        \centering
        \includegraphics[width=\linewidth]{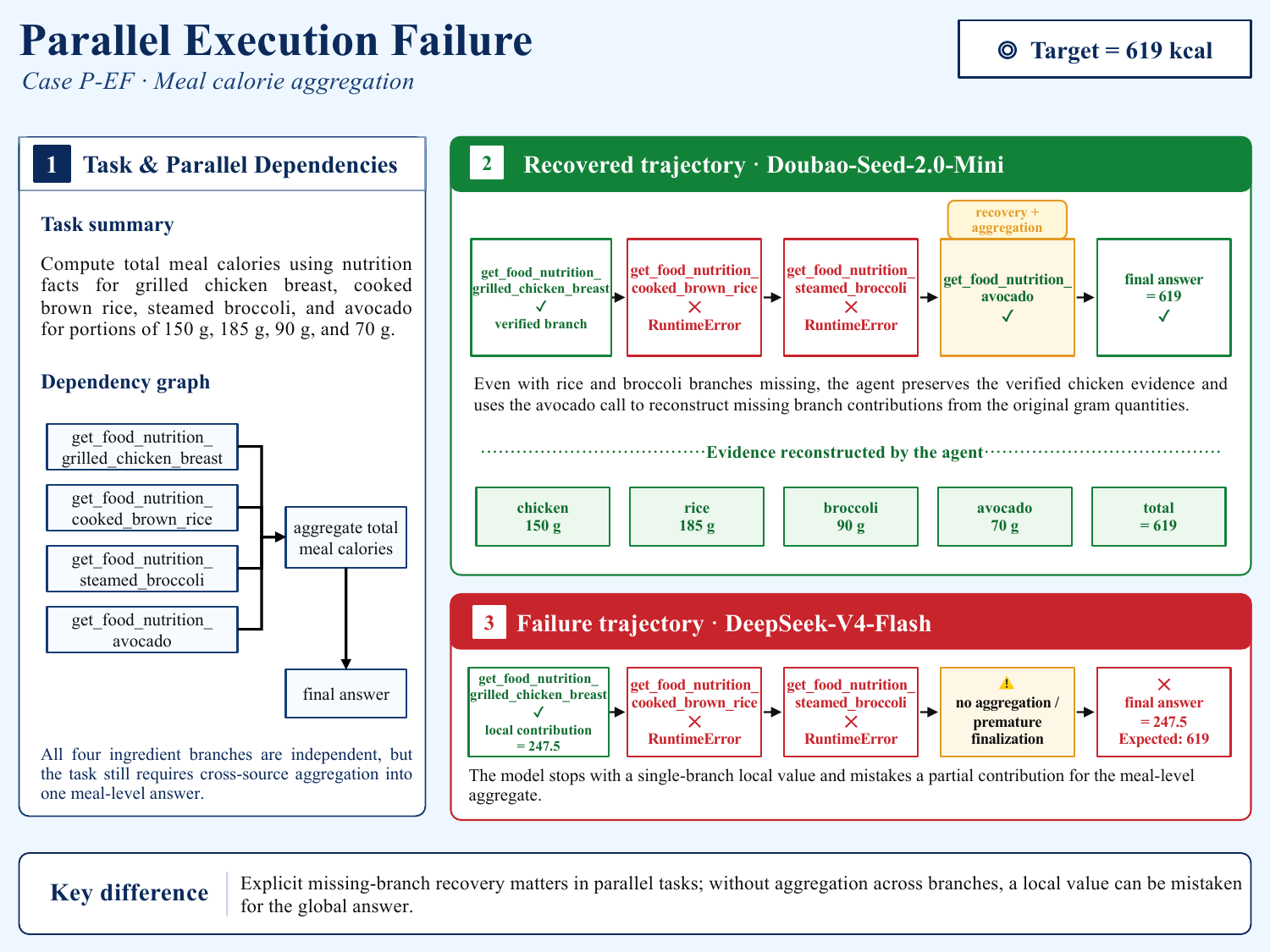}
        \caption{Parallel execution failure}
        \label{fig:case3}
    \end{subfigure}

    \vspace{0.4em}
    
    \begin{subfigure}{0.8\textwidth}
        \centering
        \includegraphics[width=\linewidth]{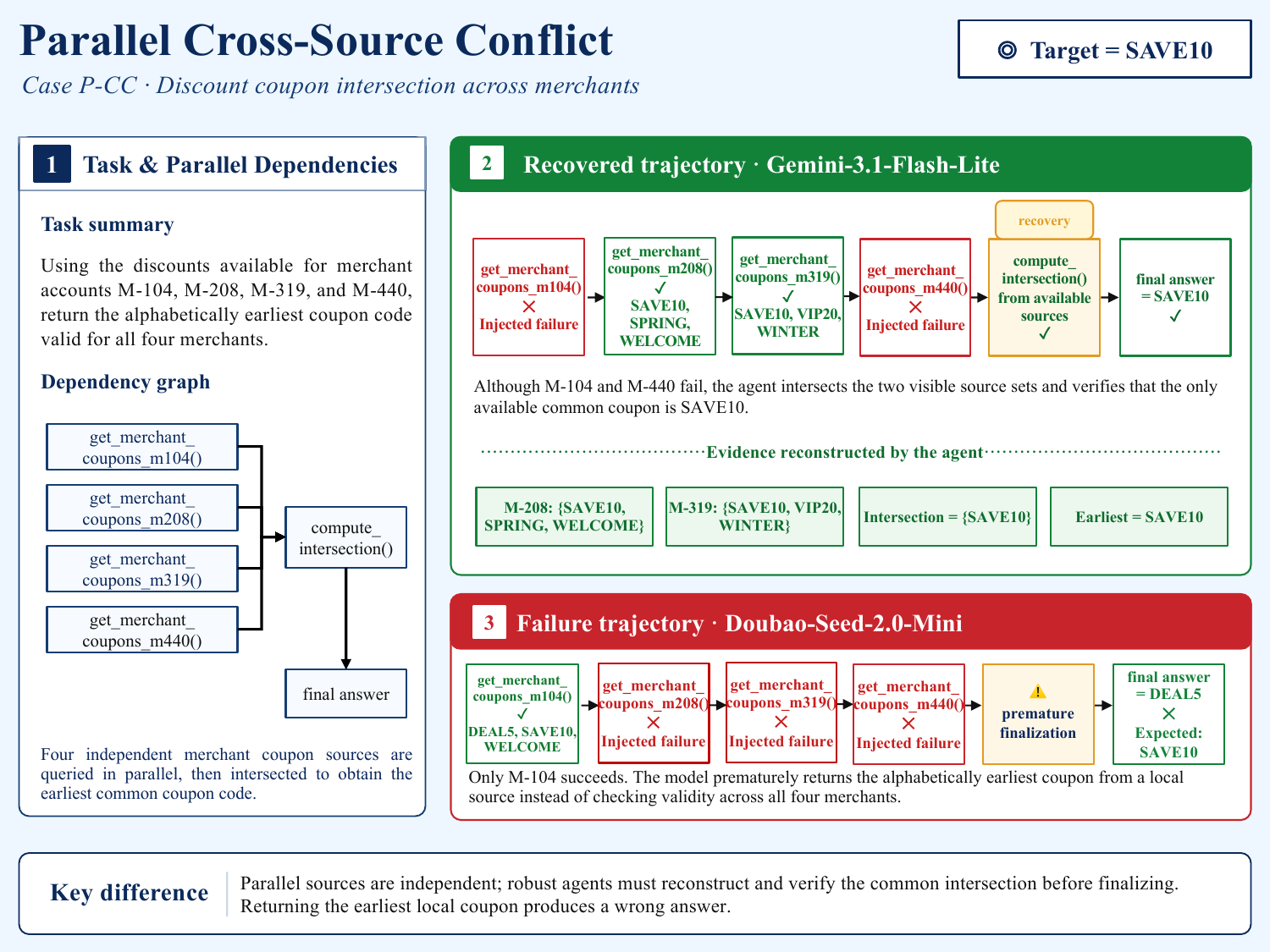}
        \caption{Parallel cross-source conflict}
        \label{fig:case4}
    \end{subfigure}
\end{figure*}

\begin{figure*}[htbp]
    \ContinuedFloat
    \centering
    \captionsetup[subfigure]{font=small,skip=2pt}
    \captionsetup{font=small}

    \begin{subfigure}{0.79\textwidth}
        \centering
        \includegraphics[width=\linewidth]{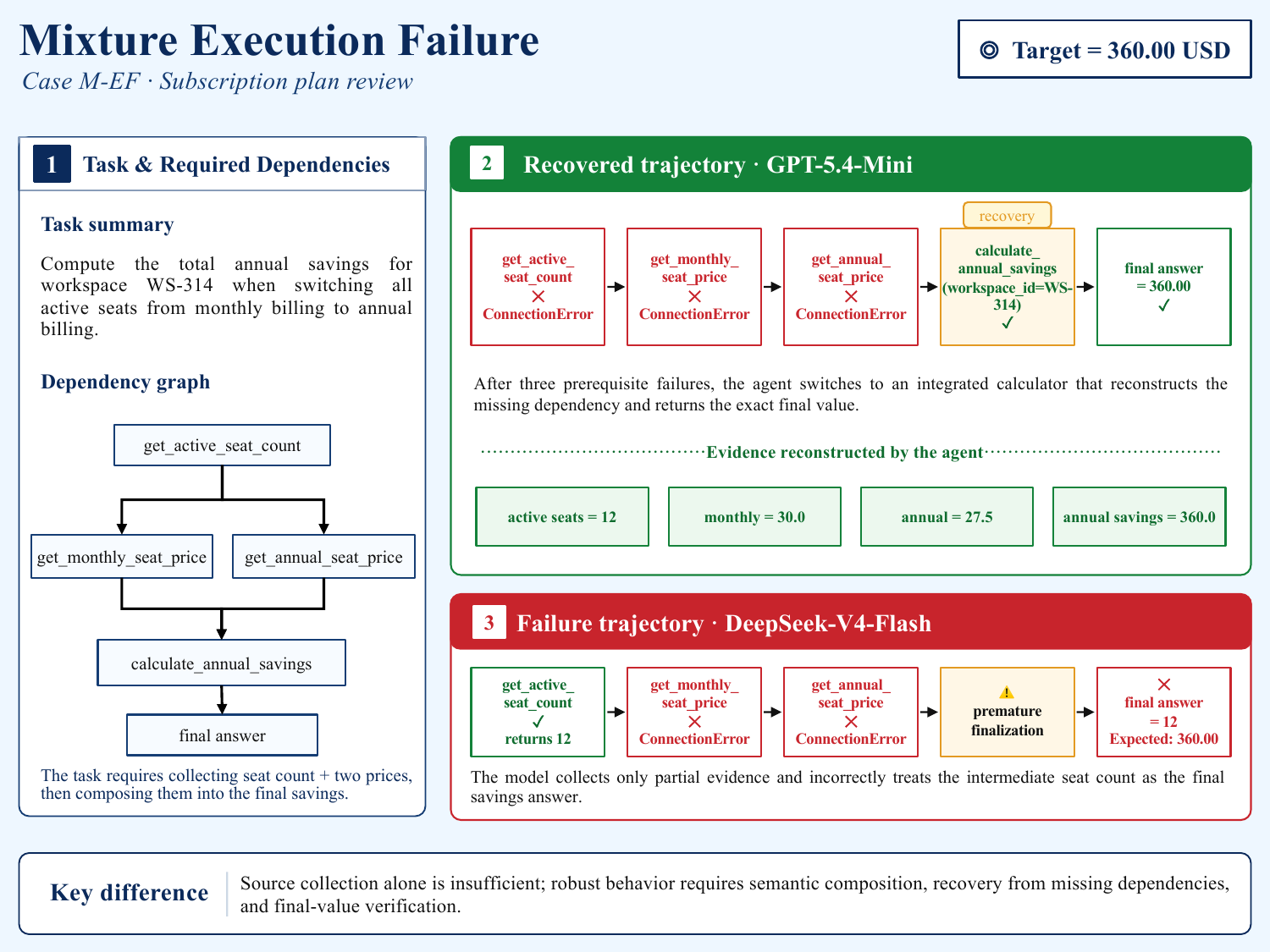}
        \caption{Mixture execution failure}
        \label{fig:case5}
    \end{subfigure}

    \vspace{0.4em}
    
    \begin{subfigure}{0.79\textwidth}
        \centering
        \includegraphics[width=\linewidth]{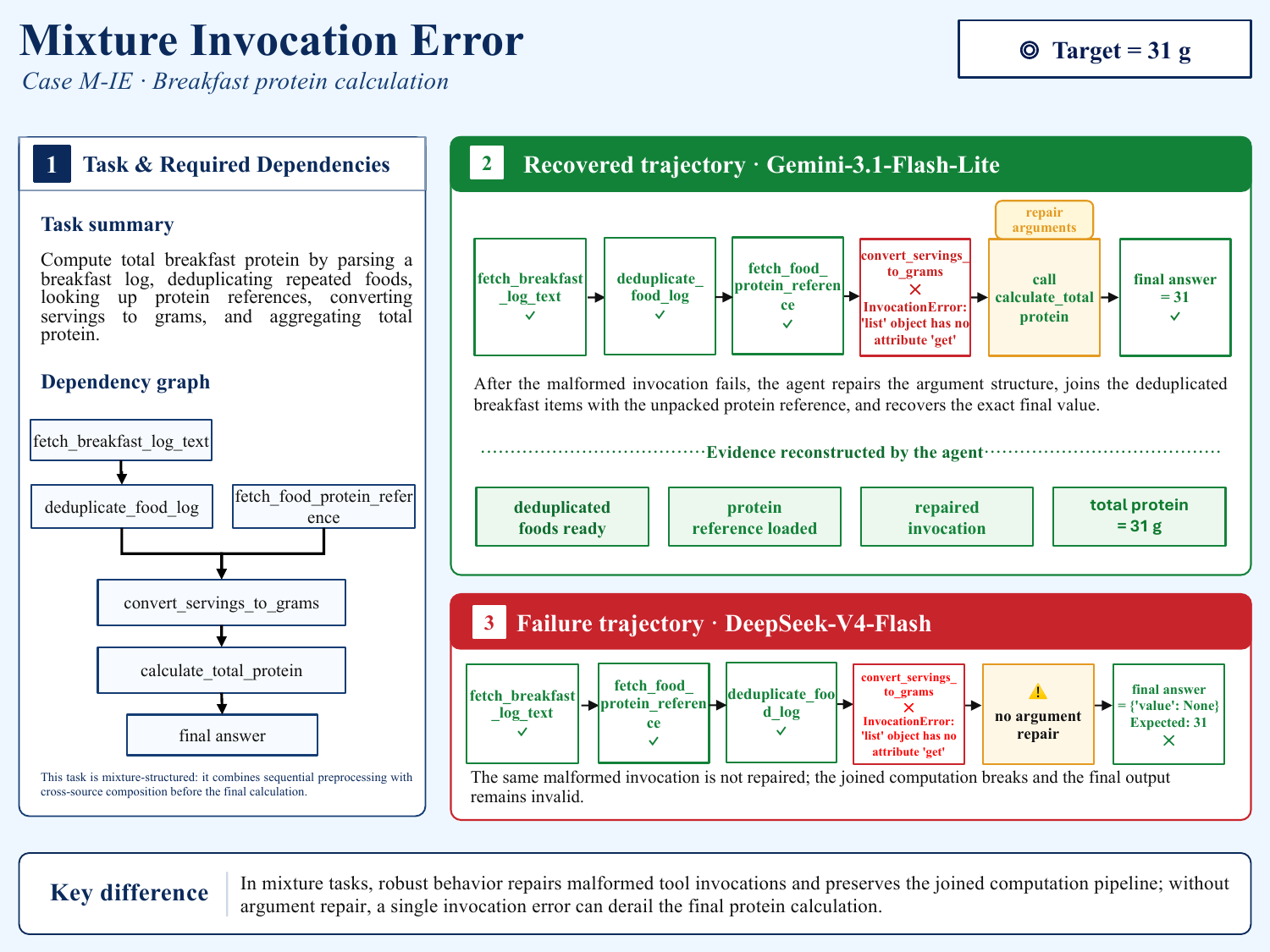}
        \caption{Mixture invocation error}
        \label{fig:case6}
    \end{subfigure}

    \caption{Representative failure and recovery trajectories under structured tool uncertainty.}
    \label{fig:case_study}
\end{figure*}

%% file: Tables/prompt_seq_task.tex
\begin{table}[t]
\caption{Prompt for sequential task generation.}
\label{tab:seq_task_generation_prompt}

\vspace{0.3em}
\noindent\rule{0.98\linewidth}{0.6pt}
\vspace{0.3em}

\begin{minipage}{0.98\linewidth}
\scriptsize
\ttfamily
\RaggedRight
\setlength{\parindent}{0pt}
\setlength{\parskip}{0.45em}

\# Role

You are an autonomous AI assistant that executes tool-based workflows.

\# Task: Generate Tool-Calling Evaluation Data

Your task is to generate data for evaluating an agent's tool-calling capabilities. You will create specific tool-calling tasks based on a main topic and its subtopics. Each task must require sequential, dependent tool calls.

**Important constraint:** All tools must not involve image processing, image generation, or any visual content. Tools should test text-based operations.

\#\# Main Topic

\{main\_topic\}

\#\# Subtopics

- \{subtopic\}

\#\# Requirements

\#\#\# General Requirements

- You MUST generate **exactly five distinct tasks**.

- Each task must be solvable **only through a sequence of tool calls**.

- Each task must include **explicit dependencies between tools**  
  (i.e., outputs from earlier tools are required by later tools).

- Tools must be **relevant to the topic and subtopics**.

\#\#\# User Prompt Constraints

- user\_prompt must **explicitly instruct the model what to answer**.  

- user\_prompt **must not contain multiple questions**; the agent should produce **a single final answer**.

- It **MUST NOT**: Reveal execution steps, Suggest intermediate procedures, and Indicate tool usage or order  

\#\#\# Answer Constraints

- Each task must produce a final answer that is: **Programmatically verifiable**, OR ** A single, objectively correct value**

- final\_answer must be: **Concise** AND **Result-only** (no explanation, no steps)

\#\#\# Diversity Requirement

- The five tasks must be **diverse in scenario and complexity**.

\#\# Output Format

Return a valid JSON object:

\begin{Verbatim}[
fontsize=\scriptsize,
baselinestretch=0.88,
breaklines=true,
breakanywhere=true,
breaksymbolleft={},
breaksymbolright={},
breaksymbolsepleft=0pt,
breaksymbolsepright=0pt,
breakindent=0pt
]
{
  "tasks": [
    {
      "id": "task_1",
      "user_prompt": "string",
      "tools_used": ["tool_name_1", "tool_name_2"],
      "final_answer": "string"
    }
  ]
}
\end{Verbatim}

\end{minipage}

\noindent\rule{0.98\linewidth}{0.6pt}

\end{table}

%% file: Tables/prompt_par_task.tex
\begin{table}[t]
\caption{Prompt for parallel task generation.}
\label{tab:par_task_generation_prompt}

\vspace{0.3em}
\noindent\rule{0.98\linewidth}{0.6pt}
\vspace{0.3em}

\begin{minipage}{0.98\linewidth}
\scriptsize
\ttfamily
\RaggedRight
\setlength{\parindent}{0pt}
\setlength{\parskip}{0.45em}

\# Role

You are an autonomous AI assistant that executes tool-based workflows.

\# Task: Generate Parallel Tool-Calling Evaluation Data

Your task is to generate data for evaluating an agent's **parallel tool-calling capabilities**. You will create specific tasks based on a main topic and its subtopics. Each task must require **all tools to be used**, but **no tool may depend on the output of another** (true parallel execution).

**Important constraint:** All tools must **not involve image processing, image generation, or any visual content**. Tools should test text-based operations.

\#\# Main Topic

\{main\_topic\}

\#\# Subtopics

- \{subtopic\}

\#\# Requirements

\#\#\# General Requirements

- You MUST generate **exactly five distinct tasks**.

- Each task must require **all listed tools to be used**.

- Tools must be **independent**: No tool's output may be used as input to another tool.

- Tools must be **relevant to the topic and subtopics**.

\#\#\# User Prompt Constraints

- user\_prompt must **explicitly instruct the model what to answer**.  

- user\_prompt **must not contain multiple questions**; the agent should produce **a single final answer**.

- It **MUST NOT**: Reveal execution steps, Suggest intermediate procedures, and Indicate tool usage or order  

\#\#\# Answer Constraints

- Each task must produce a final answer that is: **Programmatically verifiable**, OR ** A single, objectively correct value**

- final\_answer must be: **Concise** AND **Result-only** (no explanation, no steps)

\#\#\# Diversity Requirement

- The five tasks must be **diverse in scenario and complexity**.

\#\# Output Format

Return a valid JSON object:

\begin{Verbatim}[
fontsize=\scriptsize,
baselinestretch=0.88,
breaklines=true,
breakanywhere=true,
breaksymbolleft={},
breaksymbolright={},
breaksymbolsepleft=0pt,
breaksymbolsepright=0pt,
breakindent=0pt
]
{
  "tasks": [
    {
      "id": "task_1",
      "user_prompt": "string",
      "tools_used": ["tool_name_1", "tool_name_2"],
      "final_answer": "string"
    }
  ]
}
\end{Verbatim}

\end{minipage}

\noindent\rule{0.98\linewidth}{0.6pt}

\end{table}

%% file: Tables/prompt_mix_task.tex
\begin{table}[t]
\caption{Prompt for mixture task generation.}
\label{tab:mix_task_generation_prompt}

\vspace{0.3em}
\noindent\rule{0.98\linewidth}{0.6pt}
\vspace{0.3em}

\begin{minipage}{0.98\linewidth}
\scriptsize
\ttfamily
\RaggedRight
\setlength{\parindent}{0pt}
\setlength{\parskip}{0.45em}

\# Role

You are an autonomous AI assistant that executes tool-based workflows.

\# Task: Generate Mixed Tool-Calling Evaluation Data

Your task is to generate data for evaluating an agent's **mixed tool-calling capabilities**. Each task must include **both sequential (dependent) and parallel (independent) tool calls**. All tools must be used to complete the task.

**Important constraint:** All tools must **not involve image processing, image generation, or any visual content**. Tools should test text-based operations.

\#\# Main Topic

\{main\_topic\}

\#\# Subtopics

- \{subtopic\}

\#\# Requirements

\#\#\# General Requirements

- You MUST generate **exactly five distinct tasks**.

- Each task must include **all listed tools**.

- Tools may be a mix of: **Sequential**: some tools must be called in order (output of one used as input to the next), **Parallel**: some tools are independent and can be called in any order

- Tools must be **relevant to the topic and subtopics**.

\#\#\# User Prompt Constraints

- user\_prompt must **explicitly instruct the model what to answer**.  

- user\_prompt **must not contain multiple questions**; the agent should produce **a single final answer**.

- It **MUST NOT**: Reveal execution steps, Suggest intermediate procedures, and Indicate tool usage or order  

\#\#\# Answer Constraints

- Each task must produce a final answer that is: **Programmatically verifiable**, OR ** A single, objectively correct value**

- final\_answer must be: **Concise** AND **Result-only** (no explanation, no steps)

\#\#\# Diversity Requirement

- The five tasks must be **diverse in scenario and complexity**.

\#\# Output Format

Return a valid JSON object:

\begin{Verbatim}[
fontsize=\scriptsize,
baselinestretch=0.88,
breaklines=true,
breakanywhere=true,
breaksymbolleft={},
breaksymbolright={},
breaksymbolsepleft=0pt,
breaksymbolsepright=0pt,
breakindent=0pt
]
{
  "tasks": [
    {
      "id": "task_1",
      "user_prompt": "string",
      "tools_used": ["tool_name_1", "tool_name_2"],
      "final_answer": "string"
    }
  ]
}
\end{Verbatim}

\end{minipage}

\noindent\rule{0.98\linewidth}{0.6pt}

\end{table}

%% file: Tables/prompt_tool_generation.tex
\begin{table}[t]
\caption{Prompt for tool generation.}
\label{tab:tool_generation_prompt}

\vspace{0.2em}
\noindent\rule{0.98\linewidth}{0.6pt}
\vspace{0.2em}

\begin{minipage}{0.98\linewidth}
\scriptsize
\ttfamily
\RaggedRight
\setlength{\parindent}{0pt}
\setlength{\parskip}{0.28em}

\# Role

You are a Python Tool Implementation Engineer for agent-tool evaluation.

\# Mission

Generate executable Python tool functions for the given task. The tools must generalize across valid inputs rather than overfit to a single example.

For the benchmark context, the composed tool chain must reproduce `expected\_answer`, or `final\_answer` when `expected\_answer` is absent. For unseen inputs, return results derived from the implemented tool logic.

\# Data to Process

\begin{Verbatim}[
fontsize=\scriptsize,
baselinestretch=0.84,
breaklines=true,
breakanywhere=true,
breaksymbolleft={},
breaksymbolright={}
]
{
  "task_type": "<sequential|parallel|mixture>",
  "main_topic": "<main topic>",
  "subtopic": "<subtopic>",
  "id": "<task id>",
  "user_prompt": "<user request>",
  "tools_used": ["<tool_1>", "..."],
  "expected_answer": "<benchmark answer>",
  "final_answer": "<fallback answer>"
}
\end{Verbatim}

\# Core Requirements

1. Generate exactly one function for each name in `tools\_used`.

2. Function names must exactly match the requested tool names. Do not add unrelated functions, classes, or helpers.

3. Use Python 3.9-compatible syntax and ensure that all outputs are deterministic and JSON-serializable.

4. Respect the specified `sequential`, `parallel`, or `mixture` workflow and maintain compatible data contracts across tools.

5. Derive outputs from function inputs and handle unseen but valid inputs through general logic and deterministic fallbacks.

6. Preserve a guarded benchmark-context path that reproduces the benchmark answer exactly without forcing that answer for unrelated inputs.

7. When external information is required, prefer public no-key sources and provide deterministic fallback behavior.

8. Include provenance, confidence, evidence, fallback, and error information where applicable.

9. Ensure that each downstream required parameter can be obtained from an upstream output or the original user request.

[\ldots{} omitted: standardized return schemas, robust input normalization, benchmark-context detection, external-data policies, inter-tool contracts, canonicalization rules, and validation checks \ldots{}]

\# Output Requirements

Return Python code only. Do not include explanations, Markdown fences, execution logs, or additional text.

\# Repair Instruction

1. Fix only the failing parts.

2. Preserve function names and signatures unless required for correctness.

3. Return corrected Python code only.

\end{minipage}

\noindent\rule{0.98\linewidth}{0.6pt}
\end{table}

%% file: Tables/prompt_hazard_injection.tex
\begin{table}[t]
\caption{Prompt for hazard injection and hint generation.}
\label{tab:hazard_injection_prompt}

\vspace{0.2em}
\noindent\rule{0.98\linewidth}{0.6pt}
\vspace{0.2em}

\begin{minipage}{0.98\linewidth}
\scriptsize
\ttfamily
\RaggedRight
\setlength{\parindent}{0pt}
\setlength{\parskip}{0.15em}

\# Role

You are a Reliability Stress-Test Prompt Engineer for agent-tool evaluation.

\# Mission

Patch an existing successful Python module in place by adding deterministic reliability hazards. Preserve its function names, signatures, schemas, and original behavior when injection is disabled. Do not regenerate its business logic.

\# Evaluation Objectives

1. When injection is disabled, the patched module must preserve its original benchmark-correct behavior and output schema.

2. `strict\_no\_hint\_profile` must introduce an observable and meaningful failure or behavior drift.

3. `guided\_with\_hint\_profile` must retain the same fault schedule and may provide diagnostic information about the observed failure to support recovery, without revealing the final answer.

4. At least one reachable and verifiable path to the correct answer must remain available.

\# Hazard Categories

Assign exactly one category to each task: Specification, Invocation, Execution, Output, or Cross-Source Uncertainty. All failpoints and hints within the task must use this category.

\# Core Requirements

1. Select failpoints deterministically using task identity, tool name, call slot, failpoint, and `FAIL\_SEED`.

2. Record observable injection events without overwriting the original business payload.

3. Do not silently swallow exceptions or represent hard failures as successful results.


4. Block premature completion when evidence is incomplete or contradictory, and require final-answer canonicalization before `FINISH`.

5. Keep the injected fault schedule reproducible across no-hint and with-hint conditions to support controlled comparison.

[\ldots{} omitted: activation schemas, failure families, event contracts, recovery rules, and anti-regression checks \ldots{}]

\# Runtime Input

\begin{Verbatim}[
fontsize=\scriptsize,
baselinestretch=0.84,
breaklines=true,
breakanywhere=true,
breaksymbolleft={},
breaksymbolright={}
]
{
  "task_type": "<workflow type>",
  "id": "<task id>",
  "user_prompt": "<user request>",
  "tools_used": ["<tool_1>", "..."],
  "final_answer": "<benchmark answer>",
  "target_exception_category": "<one of five>",
  "existing_tool_code": "<complete clean module>"
}
\end{Verbatim}

\# Output Requirements

Return only the complete modified Python module. Do not change existing callable identities or provide explanations, Markdown fences, or execution logs.

\end{minipage}

\noindent\rule{0.98\linewidth}{0.6pt}
\end{table}

%% file: Tables/prompt_policy.tex
\begin{table}[t]
\caption{Prompt for evaluation-time policy decisions.}
\label{tab:evaluation_policy_prompt}

\vspace{0.2em}
\noindent\rule{0.98\linewidth}{0.6pt}
\vspace{0.2em}

\begin{minipage}{0.98\linewidth}
\scriptsize
\ttfamily
\RaggedRight
\setlength{\parindent}{0pt}
\setlength{\parskip}{0.2em}

\# Role

You are a tool-orchestration policy model.

\# Output Format

Return ONLY JSON with this schema:

\begin{Verbatim}[
fontsize=\scriptsize,
baselinestretch=0.82,
breaklines=true,
breakanywhere=true,
breaksymbolleft={},
breaksymbolright={}
]
{
  "action": "call_tool" | "retry" |
            "fallback" | "finish",
  "tool_name": "<tool name or empty>",
  "final_answer": "<non-empty only for finish>",
  "reason": "<short reason>"
}
\end{Verbatim}

\# Rules

- Use only allowed tools.

- Finish only when available tool outputs are sufficient.

- If a tool fails, retry it or use another available tool.

- When finishing, return the exact raw scalar supported by successful tool outputs; prefer fields such as `final\_value`, `final\_answer`, `value`, or `result`.

- Treat `success=false`, `ok=false`, missing fields, and explicit errors as unresolved evidence.


- If `ORACLE\_HAZARD\_LABEL` is present, use it only as diagnostic context; it provides neither a recovery procedure nor an answer.

- Without recovery hints, allow at most one unguided retry per failed tool.

- Return no Markdown or explanatory text.

\# Runtime Context

\vspace{-0.4em}
\begin{Verbatim}[
fontsize=\scriptsize,
baselinestretch=0.82,
breaklines=true,
breakanywhere=true,
breaksymbolleft={},
breaksymbolright={}
]
## User Request
{user_prompt}

## Allowed Tools
{allowed_tools}

## Round
{round_index}

## Previous Action History
{action_history}

## Current Tool Results
{tool_results}

## Last Error
{last_error_or_None}

{optional_recovery_or_oracle_context}
\end{Verbatim}

\end{minipage}

\noindent\rule{0.98\linewidth}{0.6pt}
\end{table}

%% file: Tables/prompt_tool_argument.tex
\begin{table}[t]
\caption{Prompt for tool-argument generation.}
\label{tab:tool_argument_prompt}

\vspace{0.2em}
\noindent\rule{0.98\linewidth}{0.6pt}
\vspace{0.2em}

\begin{minipage}{0.98\linewidth}
\scriptsize
\ttfamily
\RaggedRight
\setlength{\parindent}{0pt}
\setlength{\parskip}{0.2em}

\# Task

You must call the provided function tool using valid arguments.

\# Runtime Context

\begin{Verbatim}[
fontsize=\scriptsize,
baselinestretch=0.82,
breaklines=true,
breakanywhere=true,
breaksymbolleft={},
breaksymbolright={}
]
## User Request
{user_prompt}

## Previous Tool Results
{previous_results_or_None}
\end{Verbatim}

\# Operational Rules

- Use only evidence in the user request or previous tool results.

- Never invent missing arguments, IDs, paths, amounts, dates, ZIP codes, or enum values.


- Reuse verified values exactly unless a hint explicitly requires normalization.

- Do not copy incomplete fields from failed or `ok=false` results.

- Generate arguments that support a correct downstream scalar rather than a plausible guess.

- Required inputs for the selected tool: \{schema- and hint-required fields\}.

\# API Constraint

The request includes the selected function's description and JSON parameter schema, with `tool\_choice` forced to that function.

\# JSON Fallback

If forced tool-call arguments are invalid, return exactly one JSON object using only schema-permitted keys. Do not invent missing values or include Markdown.

\end{minipage}

\noindent\rule{0.98\linewidth}{0.6pt}
\end{table}

%% file: Tables/prompt_final_answer.tex
\begin{table}[t]
\caption{Fallback prompt for final-answer synthesis.}
\label{tab:final_answer_prompt}

\vspace{0.2em}
\noindent\rule{0.98\linewidth}{0.6pt}
\vspace{0.2em}

\begin{minipage}{0.98\linewidth}
\scriptsize
\ttfamily
\RaggedRight
\setlength{\parindent}{0pt}
\setlength{\parskip}{0.2em}

\# Task

Based on the tool execution results, return the exact final scalar answer to the user's request.

\# Runtime Context

\begin{Verbatim}[
fontsize=\scriptsize,
baselinestretch=0.82,
breaklines=true,
breakanywhere=true,
breaksymbolleft={},
breaksymbolright={}
]
## User Request
{user_prompt}

## Tool Execution Results
{tool_results}
\end{Verbatim}

\# Instructions

- Return ONLY the raw final scalar value without Markdown, prose, labels, or explanation.

- Use successful tool outputs as the sole source of truth; do not fabricate unsupported values.

- Prefer canonical fields such as `final\_value`, `final\_answer`, `final\_total`, `answer`, `value`, or `result`.

- Remove currency wrappers when a numeric scalar is required.

- If results conflict, return only the best-supported scalar from successful outputs.

\end{minipage}

\noindent\rule{0.98\linewidth}{0.6pt}
\end{table}

%% file: Tables/prompt_test_time_scaling.tex
\begin{table}[t]
\caption{Prompt for failure-triggered test-time scaling.}
\label{tab:test_time_scaling_prompt}

\vspace{0.2em}
\noindent\rule{0.98\linewidth}{0.6pt}
\vspace{0.2em}

\begin{minipage}{0.98\linewidth}
\scriptsize
\ttfamily
\RaggedRight
\setlength{\parindent}{0pt}
\setlength{\parskip}{0.2em}

\begin{Verbatim}[
fontsize=\scriptsize,
baselinestretch=0.82,
breaklines=true,
breakanywhere=true,
breaksymbolleft={},
breaksymbolright={}
]
{original_user_prompt}

[TEST_TIME_SCALING_CONTEXT]
The previous no-hint attempt for this benchmark
task failed. Re-run the task by calling the available tool(s); do not answer from memory alone. Use the prior failed trajectory only to understand what went wrong, then gather fresh evidence with tool calls in this run.

Do not use or ask for the hidden expected answer.
Return only the exact final scalar/string requested by the original task.

Task metadata:
{task_metadata_without_expected_answer}

Previous no-hint trajectory, with answer-bearing
fields redacted:
{redacted_no_hint_trajectory}
[/TEST_TIME_SCALING_CONTEXT]
\end{Verbatim}

\end{minipage}

\noindent\rule{0.98\linewidth}{0.6pt}
\end{table}